\documentclass{article}

\PassOptionsToPackage{numbers, compress}{natbib}
\usepackage[main, preprint]{neurips_2026}

\usepackage{amsmath}

\usepackage[utf8]{inputenc} 
\usepackage[T1]{fontenc}    
\usepackage{hyperref}       
\usepackage{url}            
\usepackage{booktabs}       
\usepackage{amsfonts}       
\usepackage{nicefrac}       
\usepackage{microtype}      
\usepackage{xcolor}         
\usepackage{graphicx}
\usepackage{enumitem}
\usepackage{amsfonts}
\usepackage{dsfont}
\usepackage{amssymb}
\usepackage{float}
\usepackage{wrapfig}
\usepackage{subcaption}
\usepackage{cleveref}
\usepackage{xcolor}
\usepackage{xspace}

\definecolor{darkblue}{rgb}{0, 0, 0.5}
\definecolor{darkred}{rgb}{0.6, 0, 0}
\hypersetup{colorlinks=true, citecolor=darkblue, linkcolor=darkred, urlcolor=darkblue}

\def\eg{\emph{e.g.,}\xspace}
\def\ie{\emph{i.e.,}\xspace}

\title{DynaTokens: Teaching Dynamics to Camera-Controlled Video Models at Test Time}

\author{
Ziqi Ma
\qquad
Hongqiao Chen
\qquad
Georgia Gkioxari \\
California Institute of Technology \\
}

\begin{document}

\maketitle

\begin{abstract}
Video generation must account for two sources of motion, one induced by the observer's camera path and the other caused by scene dynamics. An ideal camera-controlled video model should account for both motions: let users move the camera while evolving the scene dynamics. While current models handle camera-induced motion well in static settings, they struggle for dynamic scenes: objects are static, move incorrectly, or degrade in generation quality. We introduce DynaTokens, a lightweight set of learnable scene-specific tokens that teach dynamics to an existing camera-controlled world model. Our method is motivated by a simple asymmetry between the two sources of motion: whereas camera motion affects the generated view globally, object dynamics are spatially localized. Through cross-attention, DynaTokens trains the learnable tokens from a few example trajectories for a scene while keeping the base model frozen, and enables dynamics under new query camera paths. DynaTokens achieves a better simultaneous dynamics-camera tradeoff on VBench2 and WorldScore evaluations than LoRA, block finetuning, and specialized trainable-layer baselines. Analyses of token attention, ablations, and motion temporality suggest that matching the trainable interface to the structure of the learning target is important for effective adaptation. Project website: \url{https://glab-caltech.github.io/dynatokens/}
\end{abstract}

\section{Introduction}

Current camera-controlled video models~\citep{worldplay2025, lingbot-world, bahmani2026lyra, shen2026lyra2, wang2026worldcompass, genie2025} can generate navigable “worlds” from a image or text prompts, opening new possibilities for gaming, immersive media, and robotics.
However, despite their ability to follow camera trajectories, these models struggle with dynamics~\citep{mosaicmem2026, ma2026sightmindevaluatingstate}: the generated worlds often looks static, generate incorrect motion, or degrade in visual quality when the scene should evolve.

Recent work has shown that test-time training can adapt video foundation models to support new capabilities, including long-horizon generation~\citep{dalal2025one}, scientific discovery~\citep{yuksekgonul2026learning}, and reasoning~\citep{ttt_arc2025}. However, enabling scene dynamics in camera-controlled video models remains challenging because current architectures inject camera information directly into the generative flow process -- via PRoPE~\citep{li2025cameras}, Plücker ray embeddings~\citep{zhang2024raydiffusion}, or action conditioning~\citep{worldplay2025, genie2025} -- entangling camera control with dynamics generation. As a result, supervision on dynamics can inadvertently degrade camera control, and naive fine-tuning methods such as LoRA~\citep{hu2022lora} perform poorly. This observation motivates a new approach to model adaptation. We introduce DynaTokens, a test-time training method that selectively teaches scene dynamics while preserving the camera-control behavior.

DynaTokens builds on a simple principle: in video generation, camera control is global, influencing every patch across every frame, whereas scene dynamics can be spatially local, often confined to the objects that move or change. 
To exploit this distinction, DynaTokens introduces learnable tokens that steer the video latents toward localized dynamics without altering the global scene generation. 
By decoupling local dynamic control from global camera control, this design enables dynamic scene generation while preserving camera controllability. 

DynaTokens is lightweight: a few scene-specific tokens are added and learned while keeping the base model frozen. It can be trained from only a few samples ($\approx15$) and generalize to unseen camera trajectories, producing videos under novel camera controls. Importantly, these few training trajectories need not be high quality: they can be short, contain background artifacts or motion discontinuity. Through test-time training of the tokens, DynaTokens can overcome these limitations, generating temporally continuous videos with prompt-aligned dynamics for new trajectories.

We show quantitatively and qualitatively that DynaTokens effectively enables dynamics without compromising camera control, surpassing current capabilities of state-of-the-art camera-controlled world models on VBench2~\citep{zheng2025vbench2} dynamic-related categories and WorldScore~\citep{duan2025worldscore}. Baseline methods such as LoRA, block finetuning, or Test-Time Training layers struggle. Beyond dynamic motion categories, we show that DynaTokens can also enable physics and stylistic changes.

We provide analysis on the DynaTokens design to understand the disentangling of dynamics with camera control. We also probe the temporality aspects of dynamic tokens to better understand how they elicit dynamic behavior of base models. Our analyses on test-time training of dynamic tokens provide deeper insight into architecture and representation of world models, paving the way for future development of large-scale world models. We also illuminate how to effectively adapt these models at test time to enable new capabilities.
Additionally, DynaTokens offers a path toward video data generation. Existing efforts in collecting dynamic, camera-annotated data rely on synthetic environments, such as games~\citep{chen2026hydra}. DynaTokens, via efficient test-time training, opens up generation of camera-controlled videos in diverse settings.

Our main contributions are as follows:
\begin{itemize}[leftmargin=*]
\item We develop DynaTokens, which significantly improves dynamic scene generation for camera-controlled video models via test-time training, evaluated on VBench2 and WorldScore.
\item DynaTokens is a new architecture design which decouples dynamics from camera control, enables much better performance than alternatives like LoRA and TTT layers.
\item We identify a structural mismatch between existing adaptation methods and the underlying transformations they model: LoRA-type methods apply global parameter updates, whereas object dynamics is spatially localized. We provide analysis on how DynaTokens overcomes this limitation via a design that aligns with the learning target.
\end{itemize}
\section{Related work}

\paragraph{Camera conditioning in video models.}
Camera-controlled video models are flow-based video generation models that take in camera control, along with image and text, as conditioning signals. Camera conditioning can be achieved in 4 ways: noise warping~\citep{burgert2025gowiththeflow, shen2026lyra2} which warps the noise based on camera movement, PRoPE~\citep{worldplay2025, mosaicmem2026} which encodes camera frustum in positional embedding, action conditioning~\citep{worldplay2025, genie2025} which encodes discrete camera actions, and pl\"{u}cker ray embedding~\citep{lingbot-world, shen2026lyra2}, usually channel-wise concatenated. We design DynaTokens to minimally affect camera conditioning and selectively ``teach'' dynamics.

\paragraph{Dynamics in camera-controlled video models.}
Camera-controlled video models~\citep{worldplay2025, lingbot-world, shen2026lyra2, genie2025} struggle with dynamics due to a combination of 3D prior~\citep{bahmani2026lyra, shen2026lyra2} and the lack of dynamics in training data~\cite{ling2024dl3dv}. Methods that tackle dynamics can fall into two categories: (1) Improving existing general-purpose world models. These methods mostly focus on loosening 3D constraints in memory, such as~\citep{chen2026hydra, duan2026liveworldsimulatingoutofsightdynamics, mosaicmem2026}, which addresses out-of-frame dynamics. (2) Alternative formulations of 4D-aware generation models, such as BulletTime~\citep{wang2025bullettime} which allows both camera and time control, but is limited to short and small-range camera movements. The recent Atlas model~\cite{worldlabs2026atlas} enables camera-controlled dynamics generation, but requires multiple input views.

\paragraph{Test-time training.}
Test-time training methods train foundation models at test time to surpass base model capabilities in challenging tasks, such as the ARC challenge~\citep{ttt_arc2025}, scientific discovery~\citep{yuksekgonul2026learning}, and long video generation~\citep{dalal2025one}. Architecturally, test-time training methods either use LoRA~\citep{ttt_arc2025} or specialized architectures like Test-time Training (TTT) layers~\citep{zhang2025testtimetrainingright, dalal2025one}. We design DynaTokens, a new test-time training architecture for world models, to enable localized dynamic learning.

\section{Method}
\begin{figure}[t!]
  \centering
  \includegraphics[width=\textwidth]{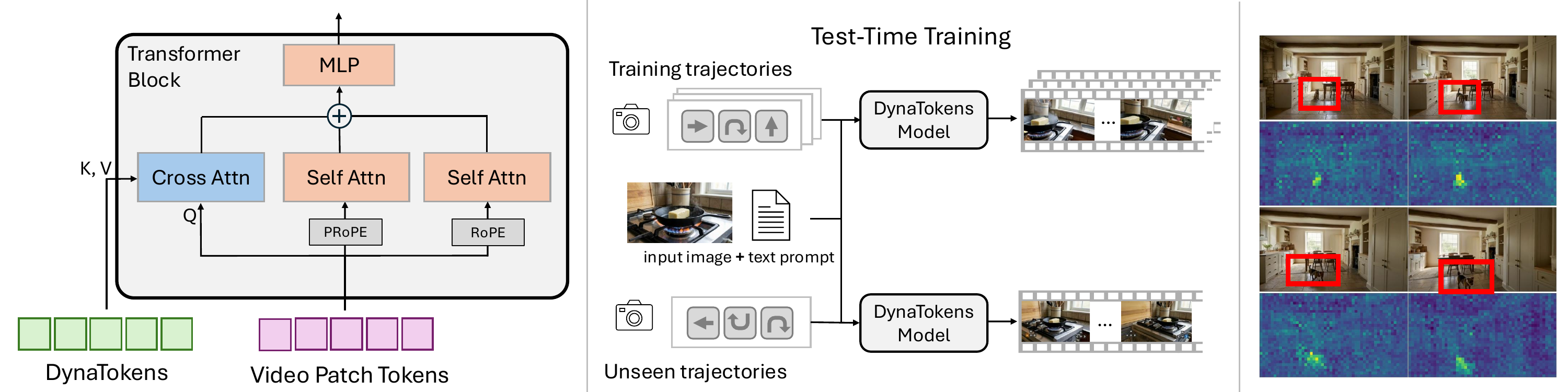}
  \begin{minipage}[t]{0.35\linewidth}
    \centering
    \small
    (a) DynaTokens design
  \end{minipage}
  \begin{minipage}[t]{0.35\linewidth}
    \centering
    \small
    (b) TTT setting
  \end{minipage}
  \begin{minipage}[t]{0.2\linewidth}
    \centering
    \small
    \hspace{0.69cm} (c) Weight maps
  \end{minipage}
  \caption{DynaTokens enables camera-controlled video models to learn dynamics via test-time training. (a) DynaTokens introduces learnable tokens that are cross-attended by the video patch tokens in every transformer block. (b) For each scene, we train DynaTokens on a small set of camera trajectories ($\approx$15), enabling inference on novel, unseen trajectories. (c) Visualization of the attention weights for a dynamic token, corroborating the localized dynamics hypothesis.}
  \label{fig:dynatoken}
  \vspace{-3mm}
\end{figure}

DynaTokens introduces scene-specific tokens into a pretrained camera-controlled video generation model. During training, the base model is kept frozen, and only the tokens are updated, enabling correct dynamics when exploring different camera trajectories for the scene. We motivate the design of dynamic tokens, then explain our architecture, test-time training setup, and training objective.

\subsection{Motivating example}
\begin{wrapfigure}{r}{0.35\textwidth}
  \vspace{-12mm}
  \centering
  \includegraphics[width=0.35\textwidth]{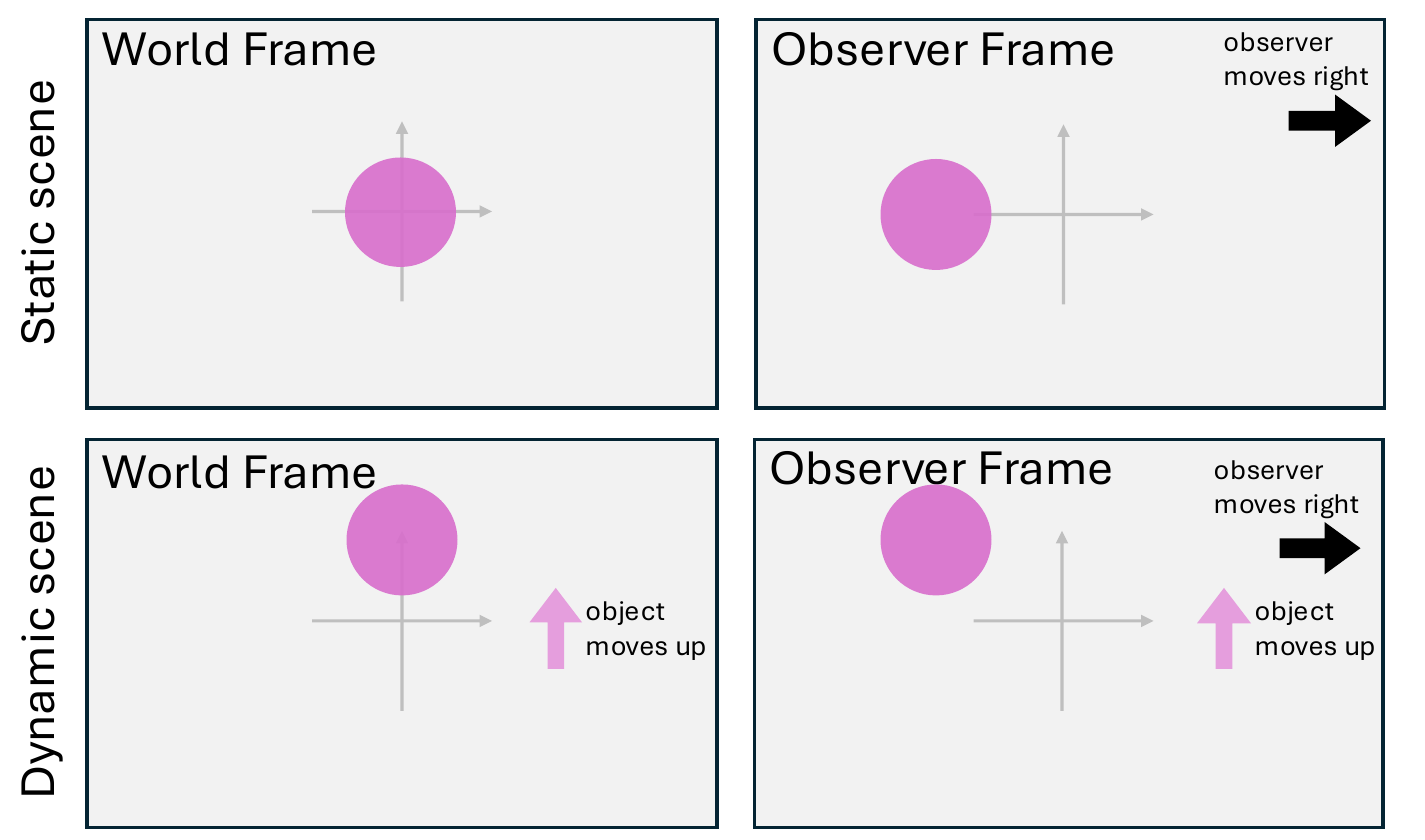}
  \caption{Motivating example.}
  \label{fig:example}
  \vspace{-6mm}
\end{wrapfigure}
Consider a simple setting where the world is a 2D coordinate frame, in which a filled unit circle, $P \subseteq \mathbb{R}^2$, is present, similar to the setting in~\citep{lillemark2026flowequivariantworldmodels}. The observer frame starts identical as the world frame.
The goal of a video generative process is to recover the world at time $t$ from the observer frame under both observer and scene motion.

\noindent \textit{Static scene, only observer moves:}
The observer moves with $\psi_t$ which represents the ``observer-to-world'' mapping: $\psi_t(\mathbf{x}_\text{obs}) = \mathbf{x}_\text{world}$ for $\mathbf{x}_\text{obs} \in \mathbb{R}^2$, where $\mathbf{x}_\text{world}$ is a point's coordinate in the world frame and $\mathbf{x}_\text{obs}$ its coordinate in the observer frame. Assume $\psi_t: (x,y) \rightarrow (x+t,y)$ (observer moves right). If the scene is static, in the observer frame, this is equivalent to $\psi_t^{-1}$ applied to object $P$: $P_{t}=\psi_t^{-1}(P) = \{(x+t,y): (x,y) \in P\}$ -- $P$ moves left.
In the observer frame, the scene is represented as $f(x,y,t) = \mathds{1}_{\{(x+t)^2+y^2 \leq 1\}}$ (top right of Fig~\ref{fig:example}).

\noindent \textit{Dynamic scene:} Consider $P$ to be a dynamic object, evolving with $\phi_t(P) \subseteq \mathbb{R}^2$. Unlike $\psi_t$ which has global effect on the whole frame, $\phi_t$ is only applied to the local subset of object points, $P$. Assume $\phi_t$ represents uniform-speed ``up'' movement: $\phi_t: (x,y) \rightarrow (x,y+t)$ in the world frame. In the observer frame, this is equivalent to $P_t = \psi_t^{-1}(\phi_t(P)) =\{(x - t,\, y + t) : (x,y) \in P\}$ -- $P$ moves left and up. 
The world from the observation frame can be represented as $f_\text{dyna}(x,y,t) = \mathds{1}_{\{(x+t)^2+(y-t)^2 \leq 1\}}$, shown in Fig~\ref{fig:example} bottom right.

\paragraph{Localized dynamics hypothesis.}
In a generative context, we care about the effect of object dynamics on the scene in the observer frame, $f$, which describes the ``static world'' and $f_\text{dyna}$ which describes the ``dynamic world''. We define operator $D$ to capture its effect, $f_\text{dyna} = Df$.

The motivating example above illustrates an important observation: at any given time, the effect of $D$ is limited to $P$'s movement. Thus, we can decompose $D$ into a dynamics function and localized gating function: $(Df)(x,y,t) = m(x,y,t)\cdot(Gf)(x,y,t)$, where $m(x,y,t)$ is a binary gating function that localizes the dynamics and $G$ applies the dynamics. In our example, $m(x,y,t)= \mathds{1}_{\{(x+t)^2+(y-t)^2 \leq 1 \lor (x+t)^2+y^2 \leq 1\}}$, and $G$ can be defined as $(Gf)(x,y,t) = f(x,y-t,t)$.

Similar to the motivating example of a circle in a plane, object dynamics in the real world are also localized: “a dog running” should affect only the dog region while preserving the rest of the scene. In other words, $D(f) = m \odot G(f)$. We build upon a base model that obeys camera control but largely keeps the scene static, \ie generates $f$, and turns it into $f_\text{dyna}$ which captures both camera and scene dynamics. We do this by recovering $D$, which we decompose into a localized gating function $m$ which identifies where dynamics should occur, and a dynamics operator $G$ that enables the motion.

\subsection{Our approach: dynamic tokens}
\label{sec:localized_gating}

To achieve this, we design a lightweight add-on architecture for pretrained camera-controlled video models, enabling test-time learning of scene dynamics while preserving camera control.
This presents two requirements: (1) the architecture needs to have enough representation capacity to ``steer'' the structure of the generated video latents while being lightweight; (2) the design should enable dynamics and camera control simultaneously.

We introduce DynaTokens, as illustrated in Fig.~\ref{fig:dynatoken}(a).
Dynamic tokens are added to transformer blocks of a video generative model. 
The input video patch tokens interact with the learned dynamic tokens via cross attention. The output is added back to the model, along with the two paths of self-attention, with and without camera control (often instantiated via PRoPE~\citep{li2025cameras}). We apply this to every double stream transformer block. The projection matrix that mixes multiple heads of cross-attention is initialized at zero to ensure minimal perturbation at initialization. We keep the base model frozen, and learn only the dynamic tokens at test time.

For each block, we introduce $K$ dynamic tokens $\mathcal{D} = \{d_j \in \mathbb{R}^m\}_{j=1}^K$. If $\mathcal{H} = \{h_i \in \mathbb{R}^n \}_{i=1}^{N}$ are the video tokens input to the block, the output of cross-attention is represented as:
\begin{equation}
h_i' = h_i + \sum_j a_{ij} \cdot v(d_j),
\qquad
a_{ij} =
\frac{\exp\!\big(q(h_i)^\top k(d_j)\big)}
{\sum_l \exp\!\big(q(h_i)^\top k(d_l)\big)}.
\end{equation}
The attention weights, $a_{ij}$, act as a gating function across the video frames, and enable application of dynamics only on relevant patches, essentially serving the role of $m$ in our localized dynamics motivation above. Since the MLP after the attention block is applied channel-wise (\ie not mixing different patch tokens), this locality is preserved across transformer blocks. We train $K=16$ dynamic tokens of dimension $m=64$ in our experiments. 

In Fig.~\ref{fig:dynatoken}(c), we visualize this gating mechanism in a real video by showing the attention weights applied on frames of a video generation with prompt: ``The dog is on the left of the table, then it runs to the front of the table''. The attention weights localize the dynamic regions, following the target object (dog) as it moves over time. We provide qualitative and quantitative analysis of attention map localization in \S\ref{sec:attention_localization} in the Appendix.

\subsection{Test-time training of dynamic tokens}

\paragraph{Setup.} Dynamic tokens are trained at test time to enable dynamic video generation. While large-scale data curation with rich scene and camera motion is difficult, DynaTokens' lightweight design enables learning only from a few videos ($\approx$15), which are easier to curate. We design a curation process leveraging VLMs and video interpolation, detailed in \S\ref{sec:curation_appendix} of the Appendix, which yields videos with dynamics. These videos might contain artifacts, and we only curate them for a few camera trajectories. Trained on these few trajectories, DynaTokens enables correct dynamics on new camera trajectories. Our setup is illustrated in Fig.~\ref{fig:dynatoken}(b).

However, sample curation alone is insufficient: fine-tuning or LoRA-adapting a pretrained camera-controlled video model on the same data yields poor results, as we show in our experiments. In contrast, DynaTokens learns to generate the right dynamics under novel, unseen camera trajectories.

\paragraph{Training objective.}

We build on pretrained camera-controlled video models. As a representative architecture, we consider autoregressive video diffusion models, which generate videos chunk-by-chunk through diffusion. We instantiate our method on HY-WorldPlay~\citep{worldplay2025}, a state-of-the-art camera-controlled video model.

Training video models is challenging: even with discretized camera actions, a single scene can admit over $10^{19}$ possible trajectories over a 5-second clip. From only a handful of training trajectories ($\approx$15) we aim to teach the add-on dynamic tokens to capture scene motion and generalize to unseen camera paths.
To reduce exposure bias, where the model conditions on its own previous-chunk predictions that drift from the training setting, we leverage a diffusion-forcing-type~\citep{chen2025diffusion} objective, similar to~\citep{worldplay2025, shen2026lyra2}:
\begin{equation}
\mathcal{L}_{\mathrm{FM}}(\theta)
=
\mathbb{E}_{l,\,t}
\left[
\left\|
v_\theta\!\big(x_l^{(t)},\, \tilde{x}_{1:l-1}^{(t)},\, t\big)
-
\big(x_l - x_l^{(0)}\big)
\right\|^2
\right]
\end{equation}
where $t$ is the flow timestep and $l$ is the chunk index. We provide ablations on our design choices and hyperparameters in \S\ref{sec:ablation}.
\section{Experiments}

We demonstrate the impact of DynaTokens: (1) whereas all current state-of-the-art camera-controlled video models struggle greatly with dynamics, DynaTokens can effectively enable dynamics within such models via test-time training; (2) alternative test-time training methods, such as LoRA~\citep{hu2022lora} or TTT layer~\citep{dalal2025one}, fail, while DynaTokens enables learning dynamics while maintaining camera control. We evaluate on a broad range of videos with dynamic changes, including motion ordering, dynamic spatial relations, text-based motion accuracy, physics, and style transformations.

\paragraph{Benchmarks and Metrics.}
Since there are no existing dynamics-focused benchmarks for camera-controlled world models, we repurpose existing video benchmarks for dynamics-related evaluation: VBench2~\citep{zheng2025vbench2} and WorldScore~\citep{duan2025worldscore}.
We evaluate two aspects: dynamics correctness and camera control correctness. 

VBench2 is a text-to-video benchmark with VLM judges. While VBench2 holistically evaluates video models, we use its dynamics-related categories, ``Dynamic Spatial Relations'' (DSR, \eg ``a dog is on the left of the table, then the dog runs to the front of the table'') and ``Motion Order Understanding'' (MOU, \eg ``a person is cooking, then they suddenly start organizing the pantry''). The original evaluation assumes a static camera, thus we make slight modifications to accommodate camera movement, such as not evaluating on frames where the main object is out-of-view due to camera movement. For camera, we calcaulte an aggregate score over rotation and translation error like WorldScore~\citep{duan2025worldscore}, leveraging state-of-the-art camera estimation method, ViPE~\citep{huang2025vipe}. We evaluate on a random subset of 20 prompts across the two categories, each for 6 different camera trajectories, on a total of 120 videos for VBench. Standard errors are reported in the Appendix.

WorldScore provides an input image and text prompt, and evaluates on optical-flow-based heuristics for dynamics. We focus on ``Motion Accuracy'' (MA) and ``Camera Control'' (Cam) scores. The original motion evaluation assumes a static camera, thus we make slight modifications to accommodate the moving camera, detailed in \S\ref{sec:metrics_details} of the Appendix. We evaluate on 15 scenes selected randomly from the benchmark, each with 6 different camera trajectories, on a total of 90 videos for WorldScore. Standard errors are reported in the Appendix.

\subsection{Comparing with state-of-the-art camera-controlled models}

\begin{figure}[h!]
  \centering
  \includegraphics[width=0.9\textwidth]{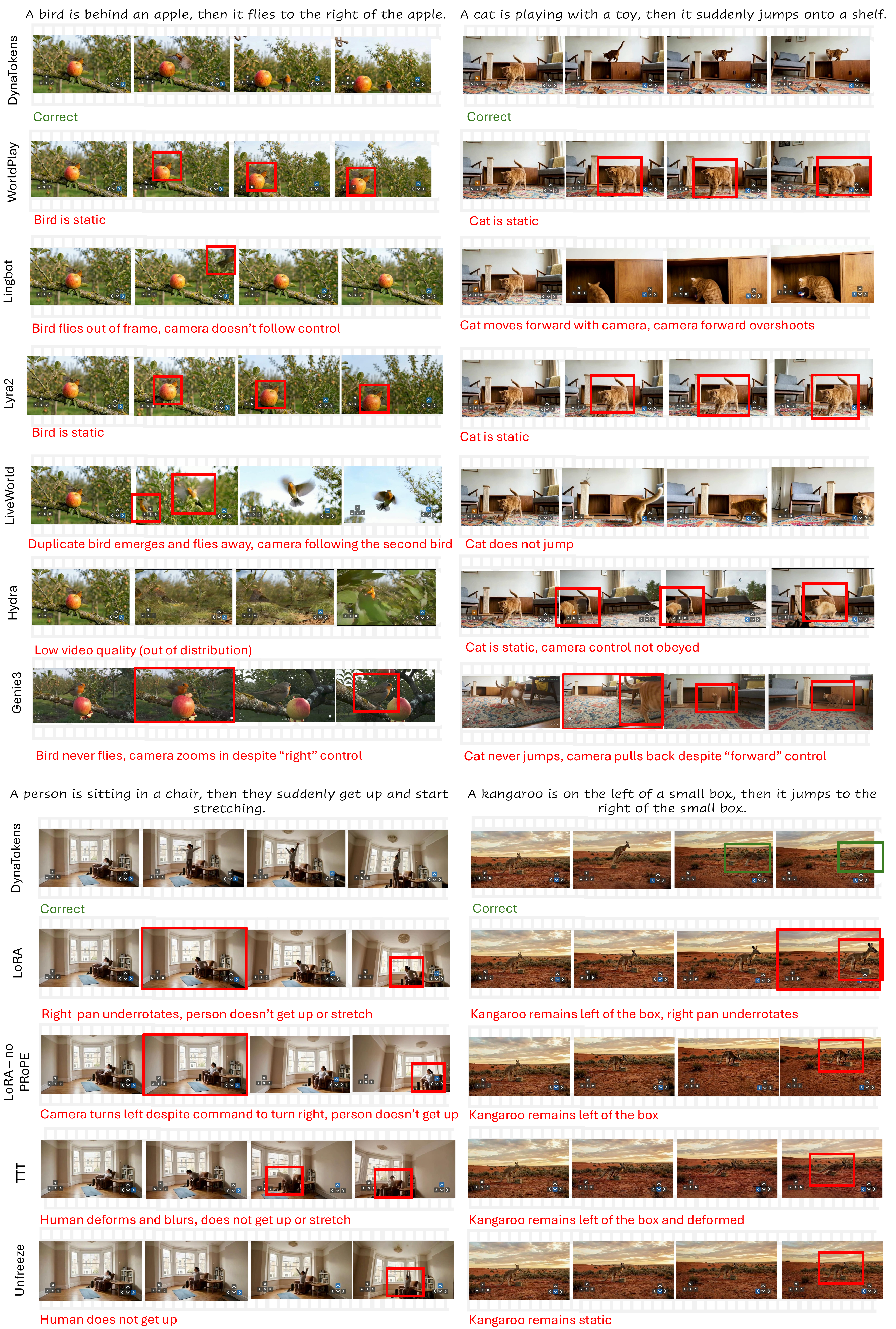}
  \caption{Top: Comparison of DynaTokens with state-of-the-art camera-controlled video models. DynaTokens consistently enables dynamics, whereas existing models struggle, either keeping the object static, show incorrect motion, or produce low-quality videos. Bottom: Comparison across different test-time training strategies. Only DynaTokens is able to correctly learn dynamics while maintaining camera control. The camera control is overlayed on the frames using keyboard schema, ``wasd'' means translation and arrows mean rotation.}
  \label{fig:qualitative}
  \vspace{-9mm}
\end{figure}

State-of-the-art camera-controlled models struggle with dynamics. While methods like Hydra~\citep{chen2026hydra} and LiveWorld~\citep{duan2026liveworldsimulatingoutofsightdynamics} improve out-of-frame dynamics by modifying their memory, they still struggle with in-frame dynamics. We show that DynaTokens, via test-time training, effectively enables dynamics.

We compare with a suite of general camera-controlled video models -- WorldPlay~\citep{worldplay2025}, Lingbot~\citep{lingbot-world}, and Lyra2~\citep{shen2026lyra2} -- as well as dynamics-focused methods -- Hydra~\citep{chen2026hydra} and LiveWorld~\citep{duan2026liveworldsimulatingoutofsightdynamics}. For WorldPlay, we evaluate their camera-post-trained checkpoint. We also compare with Genie3~\citep{genie2025} qualitatively on a small set of samples due to the UI-only access and queue time. Refer to \S\ref{sec:metrics_details} for more details. 

\begin{table*}[t!]
\begin{subtable}[b]{0.48\linewidth}
\centering
\resizebox{\linewidth}{!}{
\begin{tabular}{l|ccc|cc}
& \multicolumn{3}{c|}{VBench2} & \multicolumn{2}{c}{WorldScore} \\
  & DSR & MOU & Cam & MA & Cam \\
\midrule
Lyra2      & 0.54 & 0.17 & 0.87 & 1.11 & \textbf{0.70} \\
Lingbot    & 0.68 & 0.25 & 0.82 & 4.37 & 0.64 \\
WorldPlay  & 0.56 & 0.07 & \textbf{0.88} & 1.95 & 0.69 \\
Hydra      & 0.51 & 0.18 & 0.61 & 2.51 & 0.56 \\
LiveWorld  & 0.44 & 0.35 & 0.82 & 4.49 & \textbf{0.70} \\
DynaTokens & \textbf{0.96} & \textbf{0.69} & 0.87 & \textbf{5.33} & 0.69 \\
\bottomrule
\end{tabular}
}
\caption{}
\end{subtable}
\hfill
\begin{subtable}[b]{0.48\linewidth}
\centering
\resizebox{\linewidth}{!}{
\begin{tabular}{l|ccc|cc}
& \multicolumn{3}{c|}{VBench2} & \multicolumn{2}{c}{WorldScore} \\
  & DSR & MOU & Cam & MA & Cam \\
\midrule
DynaTokens      & \textbf{1.00} & \textbf{0.72} & \textbf{0.88} & \textbf{4.46} & \textbf{0.67} \\
LoRA           & 0.61 & 0.22 & 0.83 & 2.79 & 0.56 \\
LoRA-no PRoPE  & 0.41 & 0.06 & 0.84 & 3.62 & \textbf{0.67} \\
Fine-tuning     & 0.59 & 0.67 & 0.83 & 2.66 & 0.54 \\
TTT Layer      & 0.41 & 0.11 & \textbf{0.88} & 3.79 & 0.66\\
\bottomrule
\end{tabular}
}
\caption{}
\end{subtable}
\vspace{-2mm}
\caption{(a) Comparing DynaTokens with state-of-the-art camera-controlled video models. We evaluate Dynamic Spatial Relation (DSR) and Motion Order Understanding (MOU) on VBench2 and Motion Accuracy (MA) on WorldScore, as well as camera. Current state-of-the-art models struggle greatly with dynamics. DynaTokens, via test-time training, effectively learns dynamics while maintaining camera. (b) DynaTokens outperforms alternative test-time training methods, LoRA, finetuning, and TTT layer for dynamics. Standard errors are in Table \ref{table:baselines_se} and Table \ref{table:ablation_method_se} in Appendix.}
\label{table:baselines_ablation}
\vspace{-6mm}
\end{table*}

Table~\ref{table:baselines_ablation}(a) shows results on VBench2 ``Dynamic Spatial Relation'' (DSR), ``Motion Order Understanding'' (MOU), WorldScore ``Motion Accuracy'' (MA), and cameral control (Cam) on both benchmarks. Current models have clear failure modes, as shown in Fig.~\ref{fig:qualitative} (top): 
\begin{itemize}[nosep,leftmargin=*]
\item \textbf{Static scenes}: this is a clear failure mode for both WorldPlay and Lyra2 in almost all examples.
\item \textbf{Incorrect motion}: Lingbot tends to generate wrong motions, such as the bird flying away or the cat not jumping. Genie 3 similarly generates wrong motions.
\item \textbf{Duplicate subject}: LiveWorld splits one object into multiple, such as the 2 birds.
\item \textbf{Poor visual quality}: Hydra often outputs blurry scenes (bird example).
\end{itemize}

More qualitative examples can be found in \S\ref{sec:additional_qualitative} of the Appendix, and in the project webpage. These failures showcase that current camera-controlled models, when evaluated zero-shot, struggle with dynamics. As shown in Table~\ref{table:baselines_ablation}(a), DynaTokens improves dynamics via test-time training, resulting in a $28\%$ improvement in dynamic spatial relations, $35\%$ in motion ordering on VBench2, and a $21\%$ improvement in WorldScore motion accuracy. Additionally, DynaTokens maintains camera control on both benchmarks.

\begin{wrapfigure}{r}{0.45\textwidth}
  \vspace{-2.5mm}
  \centering
  \includegraphics[width=0.45\textwidth]{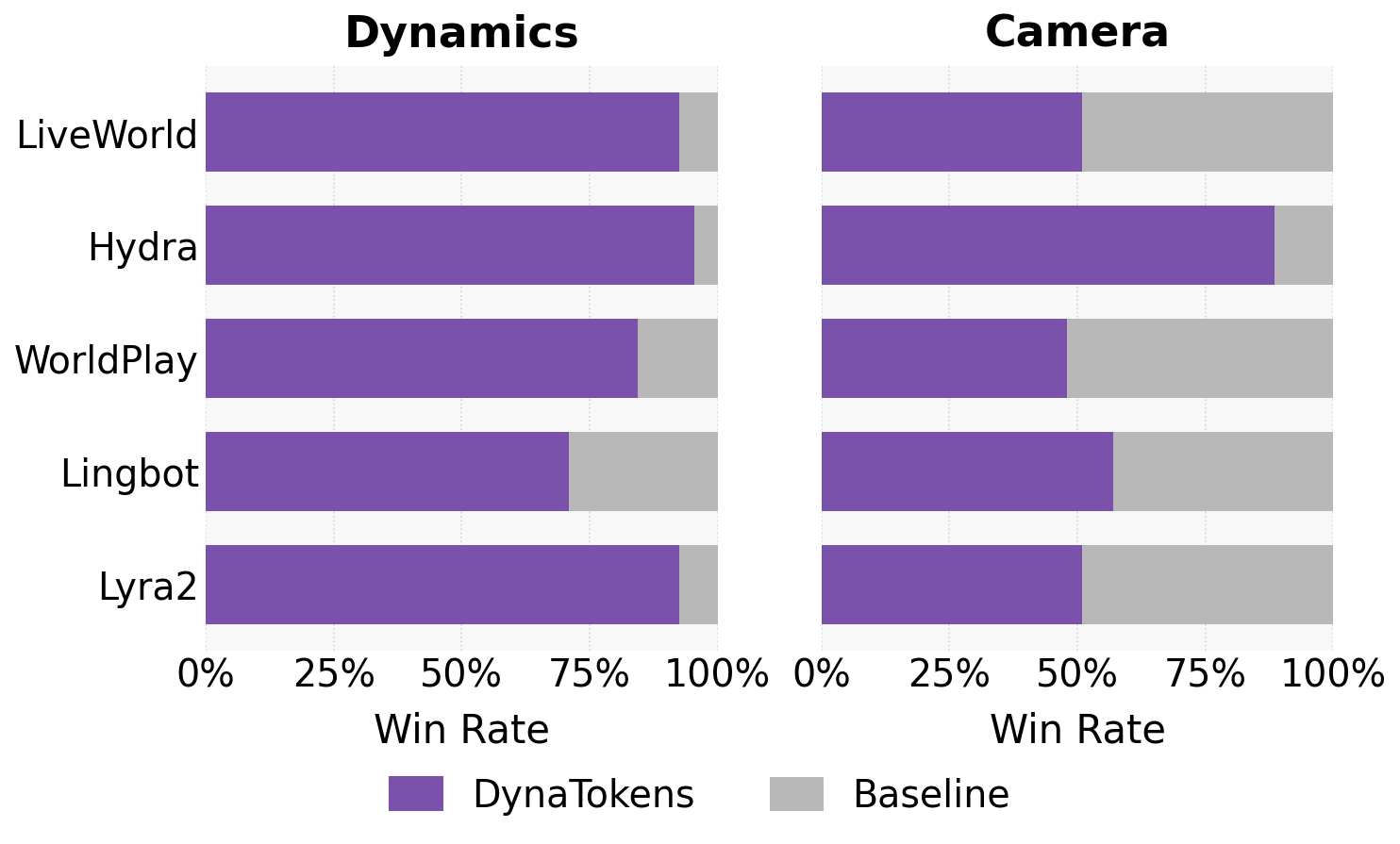}
  \vspace{-12mm}
\end{wrapfigure}
We additionally perform human evaluation, shown on the right. We perform pairwise ranking between DynaTokens and each baseline. For every baseline, we show DynaTokens' win rate against it across 100 rankings done by 5 independent annotators. DynaTokens shows clear advantage of 71\%-95.5\% win rate for dynamics. DynaTokens outperforms all other baselines except for WorldPlay on camera, with which it is a close match at $48\%$.

\subsection{Comparison with other test-time methods}

We have shown DynaTokens to effectively ``teach'' camera-controlled models dynamics while maintaining camera via test-time training in the previous section. Now, we further show that alternative test-time training methods are not as effective. We compare DynaTokens with LoRA, modified LoRA to preserve the PRoPE branch (LoRA-no PRoPE), fine-tuning last model blocks, and TTT layer for video models introduced by \cite{dalal2025one}.

Table~\ref{table:baselines_ablation}(b) shows results on VBench2 (DSR, MOU) and WorldScore. For each method, we use WorldPlay as the base model, and train on the same trajectories. We evaluate on 3 seen and 3 unseen camera trajectories for each scene to calculate the average score for dynamics and camera control. We test-time train each method on 12 distinct scenes and evaluate a total of 72 videos.

As shown in Table~\ref{table:baselines_ablation}, DynaTokens shows clear advantage in dynamics learning across both benchmarks and all metrics, corroborating our hypothesis that disentangled design for localized dynamic learning enables effective learning while maintaining camera control. In the qualitative examples shown in Fig.~\ref{fig:qualitative}(bottom), all other methods fail to fully learn dynamics, such as making the person ``get up and stretch'', or making the kangaroo ``jump to the right of the box''. More qualitative examples can be found in \S\ref{sec:additional_qualitative} in the Appendix and the project webpage.

\begin{wrapfigure}{r}{0.3\textwidth}
  \vspace{-10mm}
  \centering
  \includegraphics[width=0.3\textwidth]{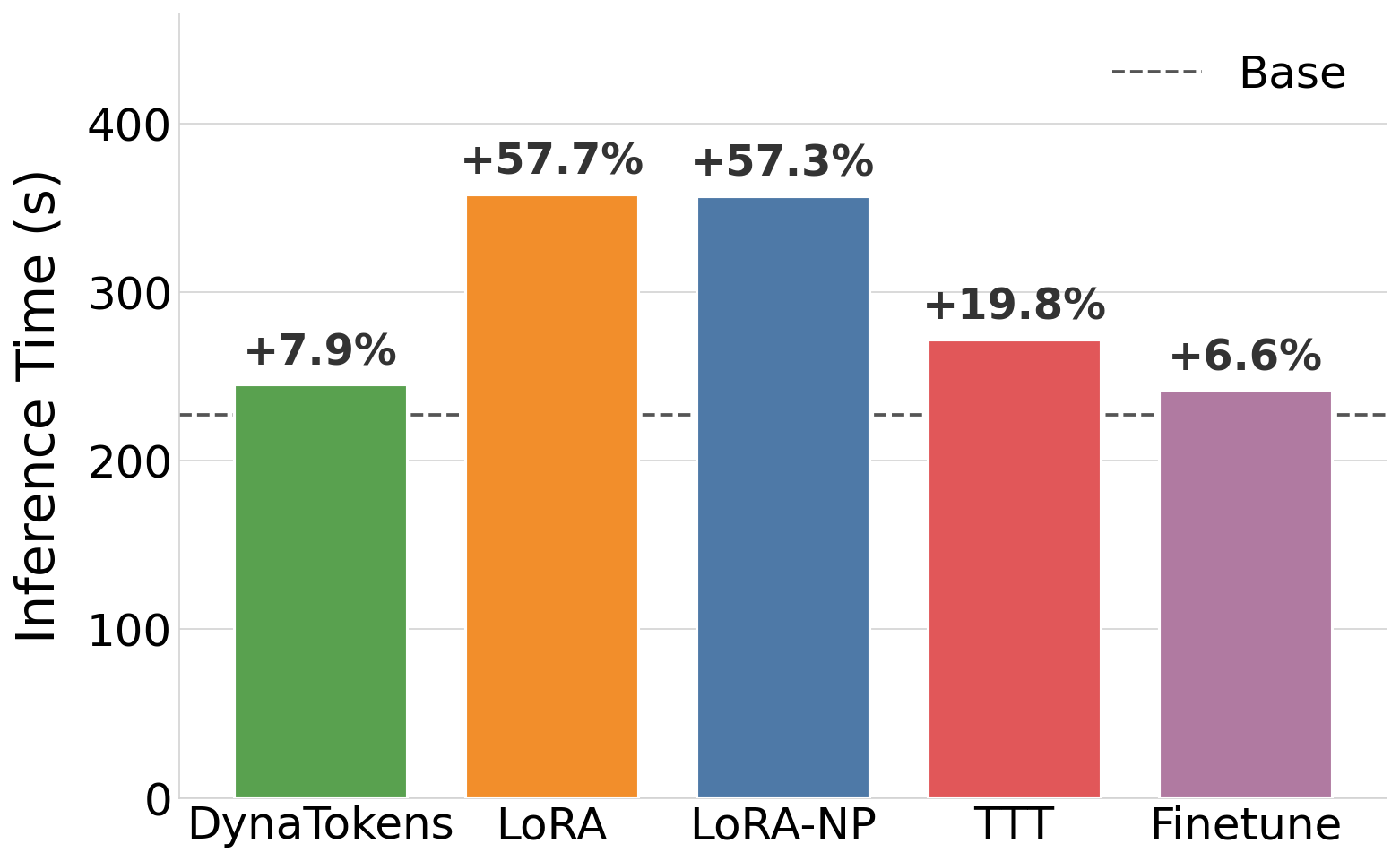}
  \vspace{-6mm}
  \caption{Inference Time}
  \label{fig:inference_time}
  \vspace{-3mm}
\end{wrapfigure}

LoRA, although high in representation capacity, cannot effectively disentangle dynamics from camera, and holistically ``steers'' the generation output, thus unable to effectively learn dynamics and yields worse camera control. We show additional analysis on LoRA and LoRA-no PRoPE in \S\ref{sec:lorafail}. Fine-tuning the last transformer blocks is also not sufficient for learning dynamics. TTT Layer, although preserving camera, shows distorted objects, as shown in the human example of Fig.~\ref{fig:qualitative}, and is unable to effectively learn dynamics. More qualitative examples can be found in the appendix, Fig.~\ref{fig:ttt_appendix1} and Fig.~\ref{fig:ttt_appendix2} in the Appendix.  We provide further analysis on the failure of LoRA and its effect on PRoPE in \S\ref{sec:lorafail}.

In addition to superior performance in both dynamics and camera, DynaTokens is also faster: as shown in Fig.~\ref{fig:inference_time}, DynaTokens only adds $7.9\%$ inference time compared to the base model (WorldPlay), while showing great improvement in dynamics. In comparison, LoRA has a much higher overhead of $57.7\%$ inference time, despite having fewer parameters (30M = 0.3\% of base model 8.6B parameters) than DynaTokens (240M, 2.7\% base model parameters). Inference is evaluated on 4 A100 GPUs.

\subsection{Generalization to Unseen Camera Paths}

\begin{wraptable}{r}{0.41\linewidth} 
\vspace{-5mm}
\centering
\resizebox{0.9\linewidth}{!}{
\begin{tabular}
{l|cc}
\toprule
 Trajectories & Dynamics & Camera \\
\midrule
Seen & 0.83 & 0.85 \\
Unseen & 0.81 & 0.86 \\
\bottomrule
\end{tabular}
}
\caption{DynaTokens demonstrates strong generalization, evaluated on VBench2.}
\vspace{-3mm}
\label{table:generalization}
\end{wraptable}
The space of all possible camera paths is large -- given the duration of videos we generate (up to 93 frames), there are more than $10^{19}$ total possible camera trajectories. While trained only trains on $\approx 15$ camera paths, DynaTokens achieves generalization to diverse camera poses. Table~\ref{table:generalization} compares DynaTokens' performance on seen and unseen camera paths on VBench2. Each evaluated unseen and seen path differ by 39.4 degrees geodesic distance on average, and differ completely in the first 28 frames' action, demonstrating sufficient spread in the exploration trajectories.

\subsection{Enabling diverse dynamics: physics and stylistic change}

We evaluate whether DynaTokens can handle diverse dynamics such as physics and style transformations. While DynaTokens is designed to learn localized dynamics, we stress-test our approach for global style transformations to probe its capability.

\begin{figure}[h]
  \vspace{-3mm}
  \centering
  \includegraphics[width=0.95\textwidth]{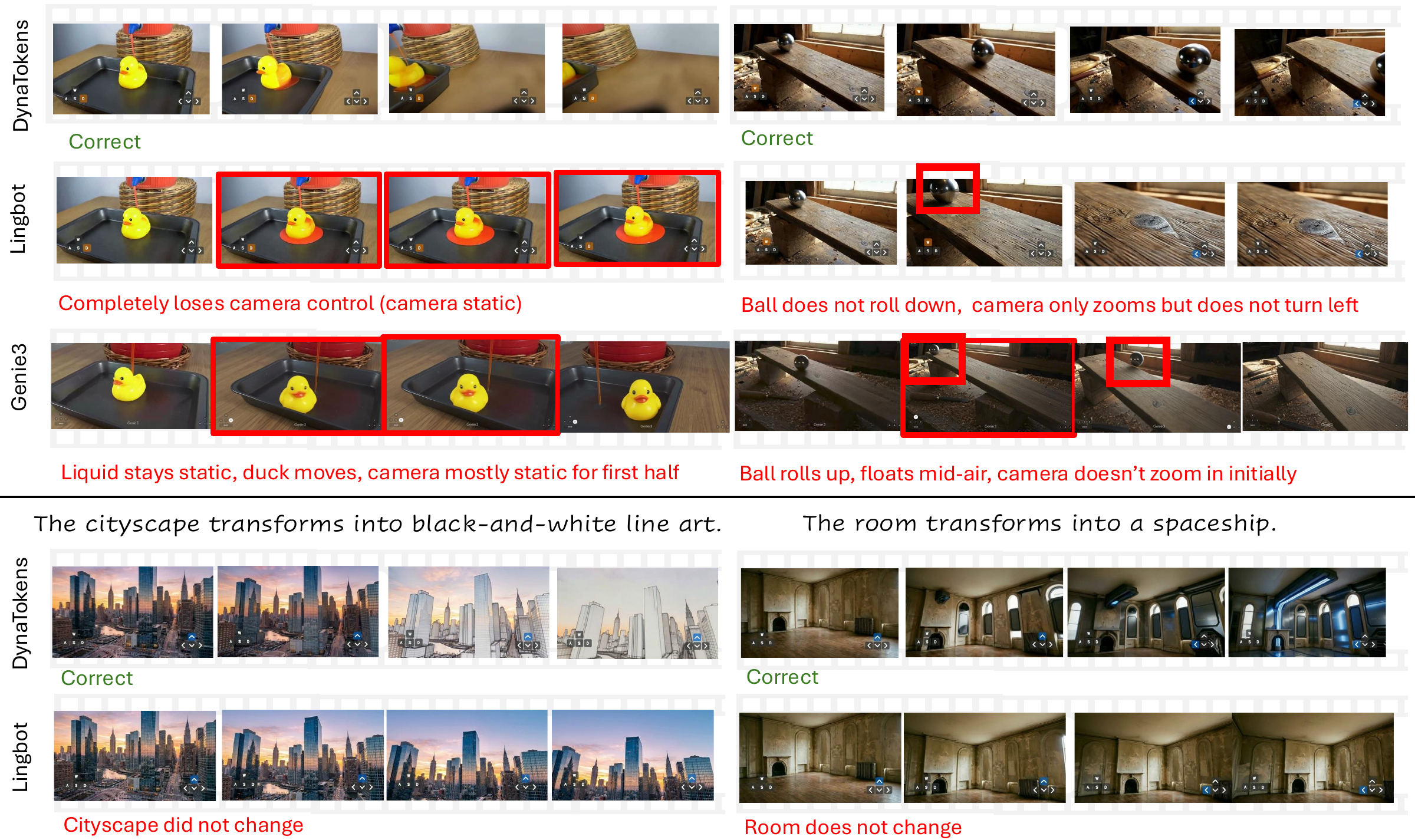}
  \vspace{-1mm}
  \caption{DynaTokens enables diverse dynamics, including physics and stylistic changes, whereas current state-of-the-art camera-controlled video models like Lingbot and Genie 3 struggle.}
  \label{fig:qualitative_physics}
  \vspace{-3mm}
\end{figure}

We show qualitative examples on Physics IQ~\citep{motamed2025physicsiq} and StevoBench~\citep{ma2026sightmindevaluatingstate}.
As shown on the top of Fig.~\ref{fig:qualitative_physics}, the best open-source state-of-the-art model, Lingbot, either fails to evolve physics (ball not rolling) or completely fails to turn the camera. Genie 3, a closed-source state-of-the-art camera-controlled model, also fails to evolve physics correctly. DynaTokens successfully learns to evolve physics in these settings. We evaluate qualitatively since Physics IQ's evaluation is based on pixel-wise alignment (IoU, MSE) under a static-camera ground truth, which cannot evaluate camera-controlled models. More examples can be found in Fig.~\ref{fig:physappendix} in the Appendix.

As shown on the bottom of Fig.~\ref{fig:qualitative_physics}, DynaTokens is capable of stylistic changes such as changing a cityscape into black-and-white line art, or transforming a room into a spaceship, while allowing the viewer to navigate the scene with camera control. Lingbot fails to transform the scene in both cases.

\section{Ablations and Analyses}
\label{sec:analyses}

\paragraph{Why does LoRA fail to capture dynamics?}
\label{sec:lorafail}

Table~\ref{table:baselines_ablation} shows that LoRA fails to simultaneously enable dynamics and preserve camera. We provide further analysis.

(1) Non-localized effect. While DynaTokens enables localized gating, LoRA is unable to represent such localized functions since its update is ``global'' to all patches: $F(h_i) = (W + BA^\top)\, h_i$.
The matrix multiplication formulation results in ``patch mixing'', and introduces global changes that interfere with camera.
(2) Preserving PRoPE. For models that use PRoPE, the QKV matrix updates from LoRA result in a shift in the PRoPE product, as detailed in \S\ref{sec:lora_appendix} of the Appendix. To alleviate this, we also experiment with LoRA-no PRoPE, which removes LoRA on the PRoPE path. However, LoRA-no PRoPE drastically worsens dynamics while only slightly improving camera compared to LoRA, as shown in Table~\ref{table:baselines_ablation}(b). 

\paragraph{Ablating the design of DynaTokens.}
\label{sec:ablation}
We ablate design of DynaTokens, and evaluate alternatives such as adding dynamic tokens directly to the patch tokens (additive), only using one token, changing the token dimensions, and eliminating context noising (using clean historical frames) during training.

\begin{wraptable}{r}{0.5\linewidth} 
\vspace{-3mm}
\centering
\resizebox{0.9\linewidth}{!}{
\begin{tabular}
{l|cccc|c}
\toprule
  & DSR & MOU  & Cam \\
\midrule
DynaTokens & \textbf{1.00} & 0.72 & 0.88 \\
Single token & 0.57 &  0.56  & \textbf{0.89} \\
Reduce token dim (32)  & 0.65 &  0.72 & 0.88 \\
Increase token dim (128) & 0.60 & 0.72 & 0.87 \\
No context noising & 0.67 & \textbf{0.83} & 0.84 \\
Additive & 0.91 & 0.67 & 0.85 \\
\bottomrule
\end{tabular}
}
\vspace{-1mm}
\caption{Ablations on dynamic token design evaluated on VBench2. The current design yields best holistic performance on dynamics and camera. DSR: Dynamic Spatial Relation. MOU: Motion Order Understanding.}
\label{table:ablation_design}
\vspace{-3mm}
\end{wraptable}
Table~\ref{table:ablation_design} shows that the current design achieves the best performance overall on VBench2 dynamics tasks, considering both dynamics and camera metrics.
Training with a single dynamic token, while slightly better on camera, compromises dynamics metrics greatly. This is likely because the softmax attention weight, which acts as localized gating, is no longer meaningful. No context noising, although slightly better on the simpler Motion Order Understanding tasks, causes a significant drop in Dynamic Spatial Relations, which require later chunks of the video to produce spatially-correct motions despite the discrepancy between the model's self prediction at inference time with teacher forcing training.

\paragraph{Probing the temporal aspect of DynaTokens.}
\label{sec:temporality}

\begin{wrapfigure}{r}{0.4\textwidth}
\vspace{-6mm}
  \centering
  \includegraphics[width=0.38\textwidth]{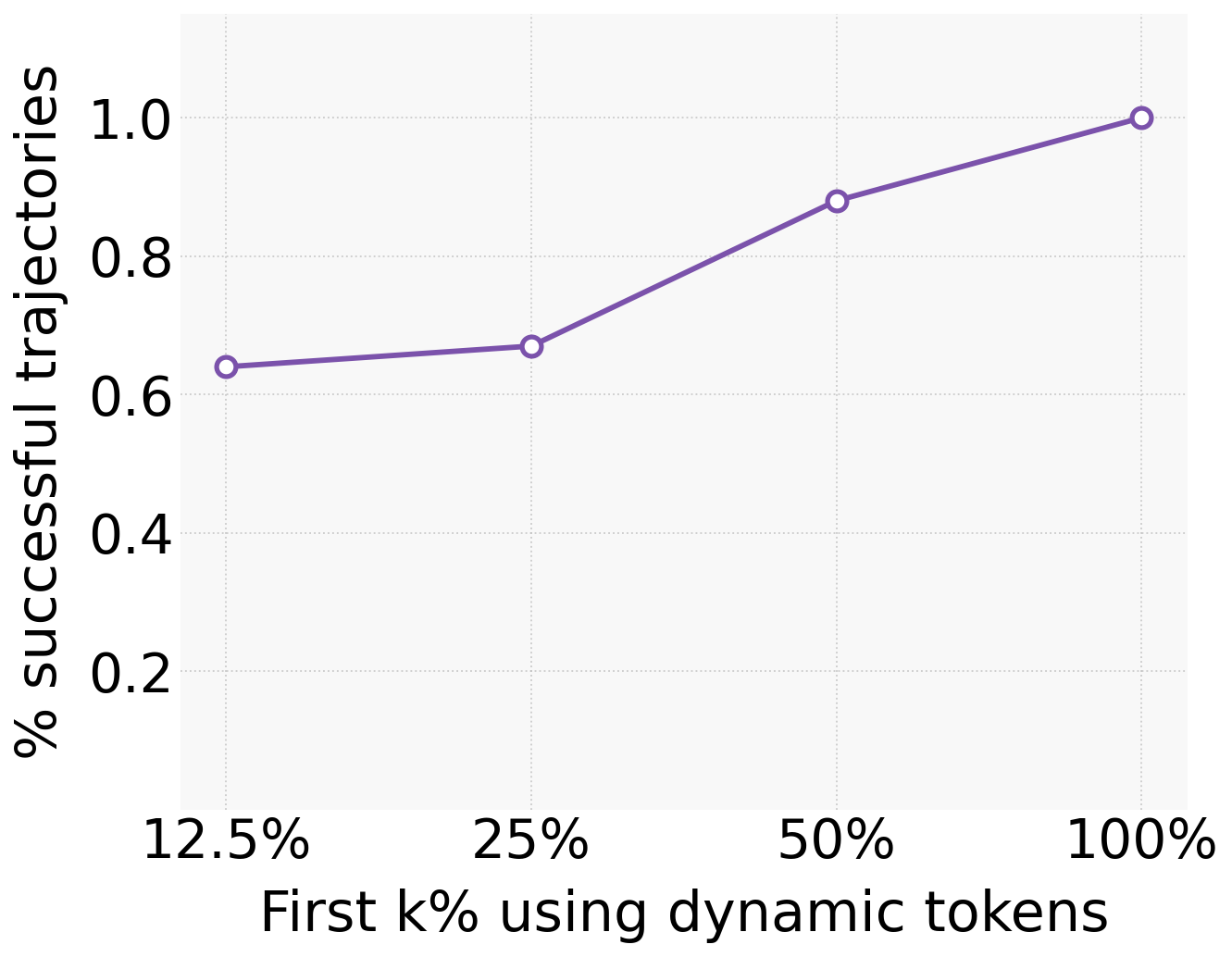}
  \vspace{-8mm}
\end{wrapfigure}
We probe whether the base model is capable of continuing dynamics after the motion is kick-started by only applying DynaTokens to the first 1/8, 1/4 and 1/2 of the full video. As shown on the right, even only using dynamic tokens for the first 1/8 can yield over 64\% successful dynamic trajectories, and using it for the first half yields successful dynamics 88\% of the time. This suggests that once the motion is ``kick-started'', the base model has the capability of continuing it via autoregressive chunk-by-chunk generation.

We also evaluate the application DynaTokens on longer videos, showcasing that the learned dynamics is not limited to the training duration, shown in Fig.~\ref{fig:longvideo_appendix} in the Appendix.

\section{Conclusions and Discussions}
Dynamic scenes are a key challenge in camera-controlled world models. We present DynaTokens, a lightweight add-on module that can be test-time-trained to enable camera-controlled video models to learn localized dynamics for a given scene. This design is able to ``selectively'' teach dynamics while enabling good camera control, whereas other designs such as LoRA fail. We provide analysis on why the design of dynamic tokens can effectively disentangle local dynamics with global camera control, and show analyses with attention weights to demonstrate the localized gating effect of our design. DynaTokens, via test-time training, greatly surpasses current state-of-the-art camera-controlled video models on VBench2 dynamic categories and WorldScore, and can extend to physical dynamics and style transfer. Our analysis on the dynamic token design sheds light on the representation of dynamics and camera control for video models. DynaTokens also opens up data generation for dynamic-rich, camera-controlled videos in diverse settings, paving the way for generating dynamic, navigable worlds in the future. For more info, visit our project page \url{https://glab-caltech.github.io/dynatokens/}.

\clearpage
\section{Ackowledgements}
We thank Damiano Marsili, Aadarsh Sahoo, Jiacheng Liu, and Jhan Liufu for discussions. This project was funded by the Packard Fellowship, the Powell Foundation, an Amazon Research Award and a Google Faculty Award. We also thank Google for providing us with Gemini credits.

\bibliographystyle{plainnat}
\bibliography{main}
\medskip

\clearpage
\appendix
\section{Additional Qualitative Results}
\label{sec:additional_qualitative}

\begin{figure}[h]
  \centering
  \includegraphics[width=\textwidth]{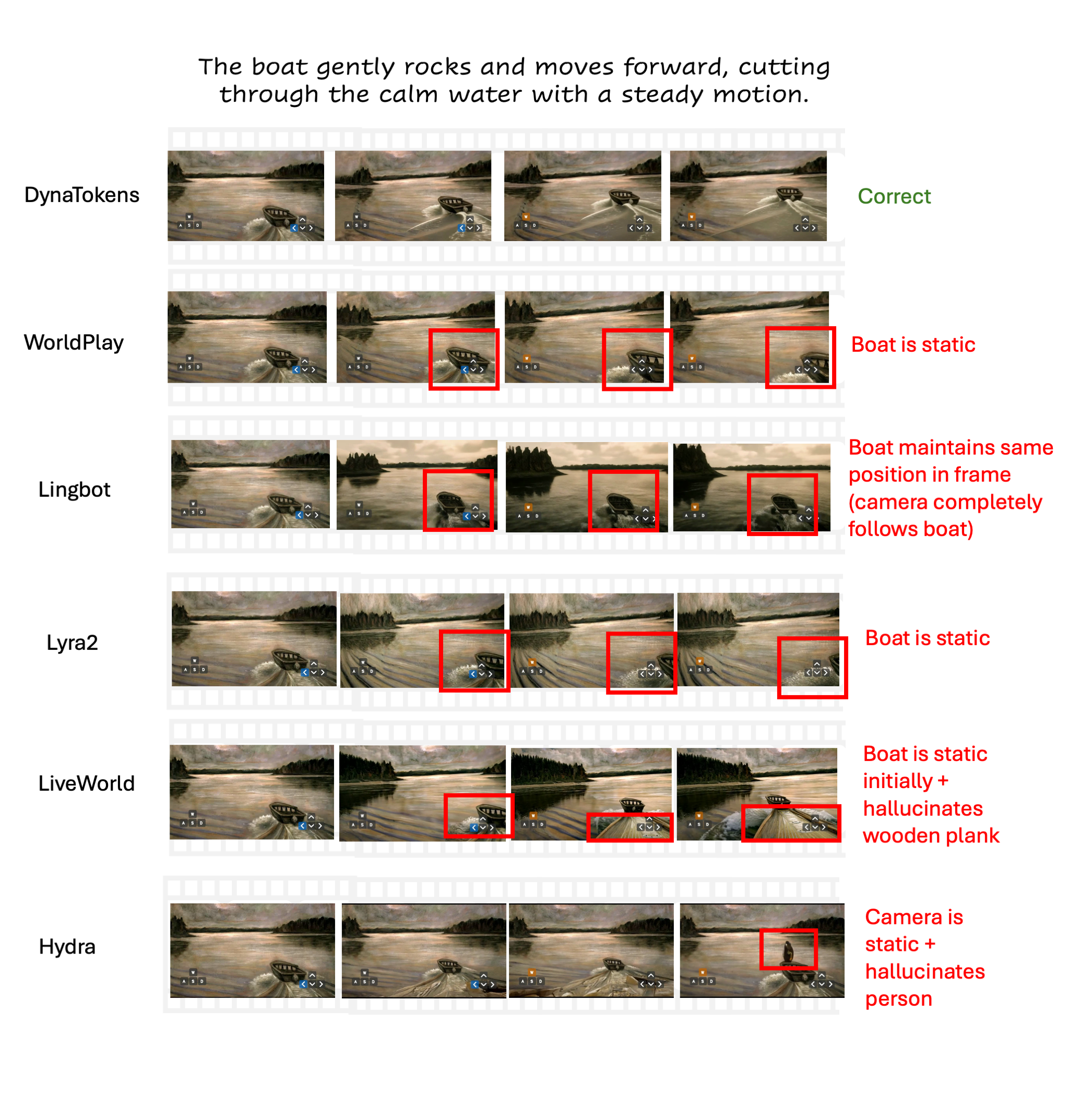}
  \caption{Additional qualitative examples of comparison with state-of-the-art camera-controlled video models.}
  \label{fig:baselines_appendix1}
\end{figure}

\begin{figure}[h]
  \centering
  \includegraphics[width=\textwidth]{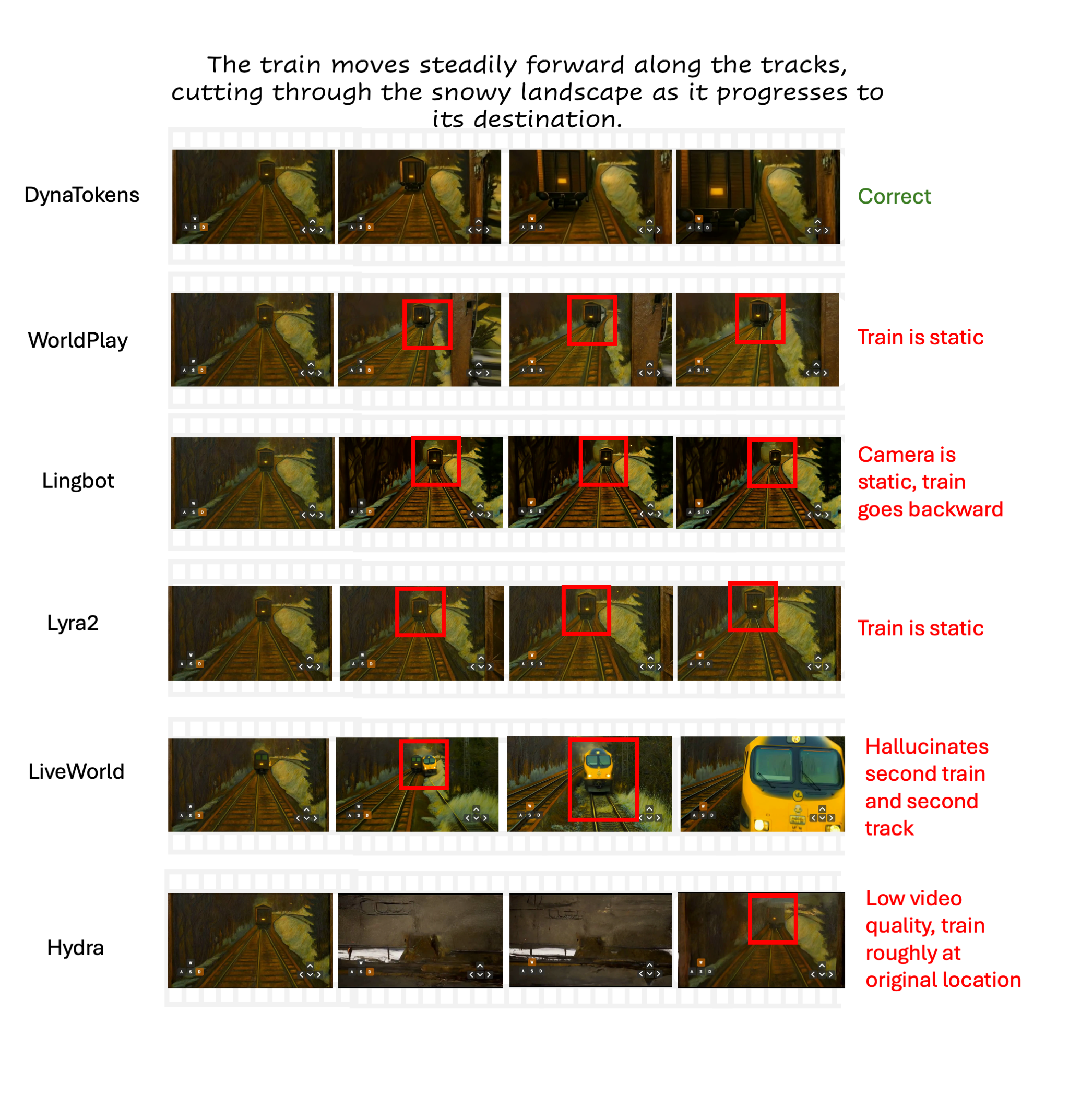}
  \caption{Additional qualitative examples of comparison with state-of-the-art camera-controlled video models.}
  \label{fig:baselines_appendix2}
\end{figure}

\begin{figure}[h]
  \centering
  \includegraphics[width=\textwidth]{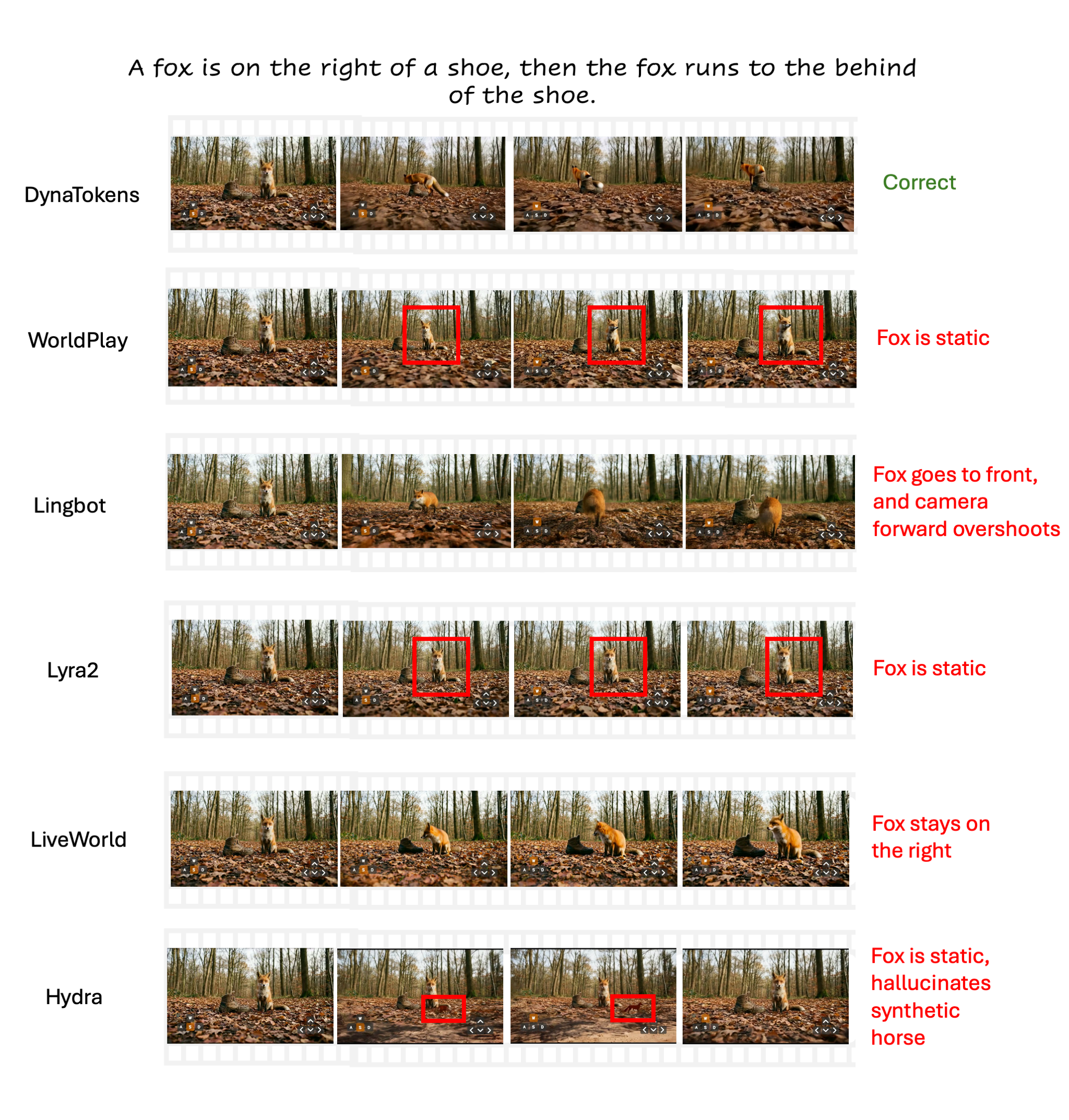}
  \caption{Additional qualitative examples of comparison with state-of-the-art camera-controlled video models.}
  \label{fig:baselines_appendix3}
\end{figure}

\begin{figure}[h]
  \centering
  \includegraphics[width=\textwidth]{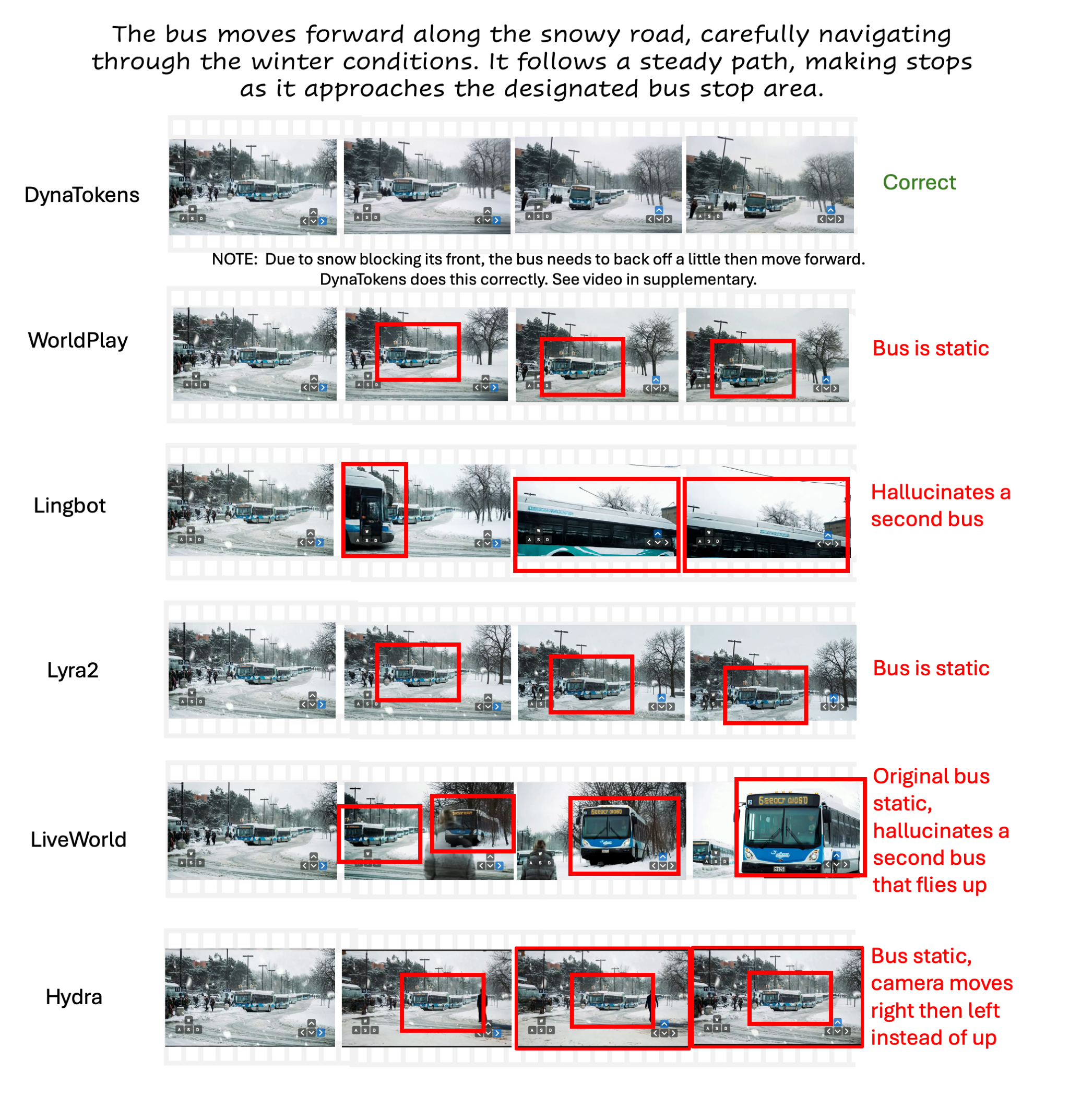}
  \caption{Additional qualitative examples of comparison with state-of-the-art camera-controlled video models.}
  \label{fig:baselines_appendix4}
\end{figure}

\begin{figure}[h]
  \centering
  \includegraphics[width=\textwidth]{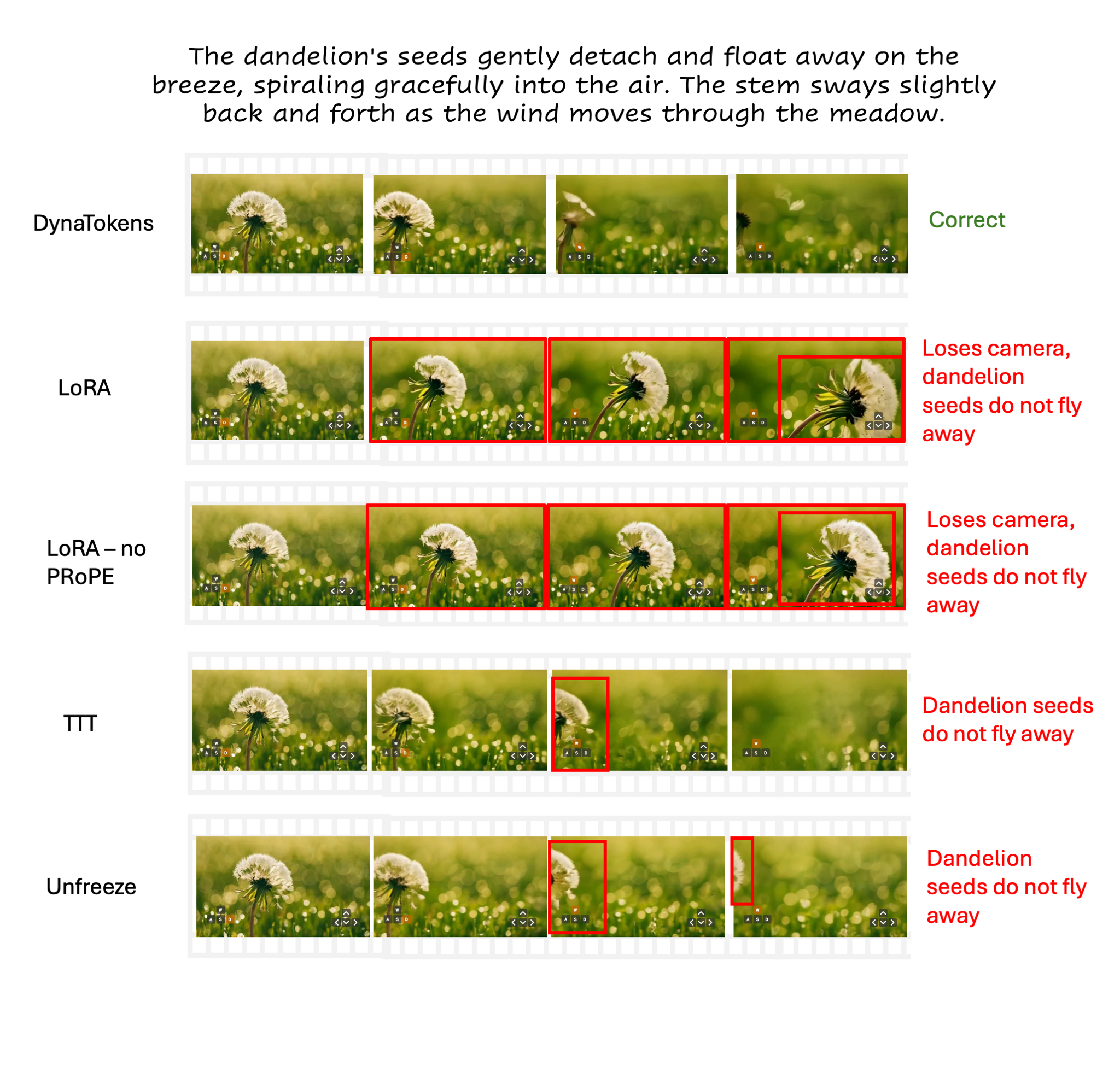}
  \caption{Additional qualitative examples of comparison across test-time methods.}
  \label{fig:ttt_appendix1}
\end{figure}

\begin{figure}[h]
  \centering
  \includegraphics[width=\textwidth]{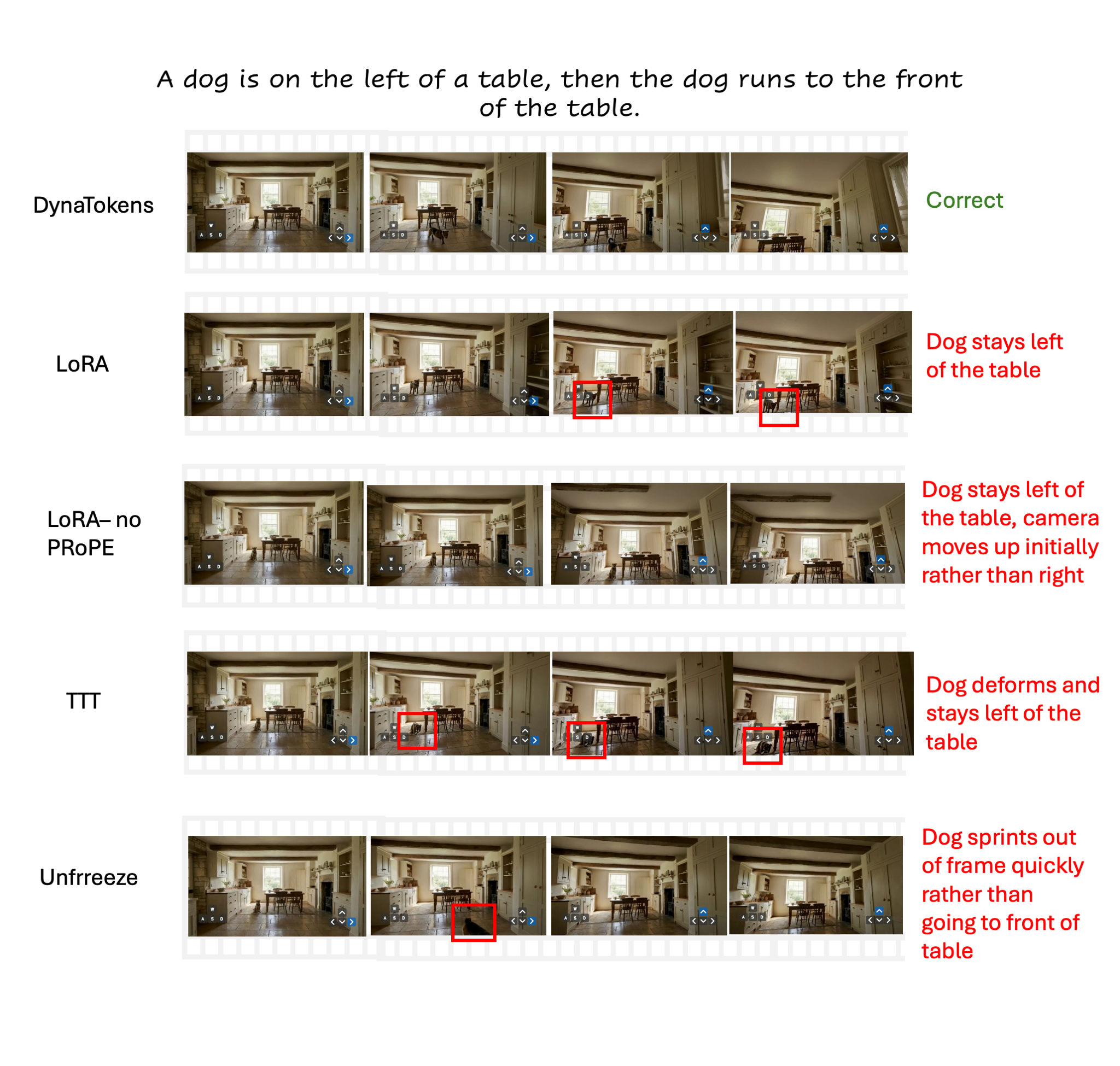}
  \caption{Additional qualitative examples of comparison across test-time methods.}
  \label{fig:ttt_appendix2}
\end{figure}

\begin{figure}[h]
  \centering
  \includegraphics[width=\textwidth]{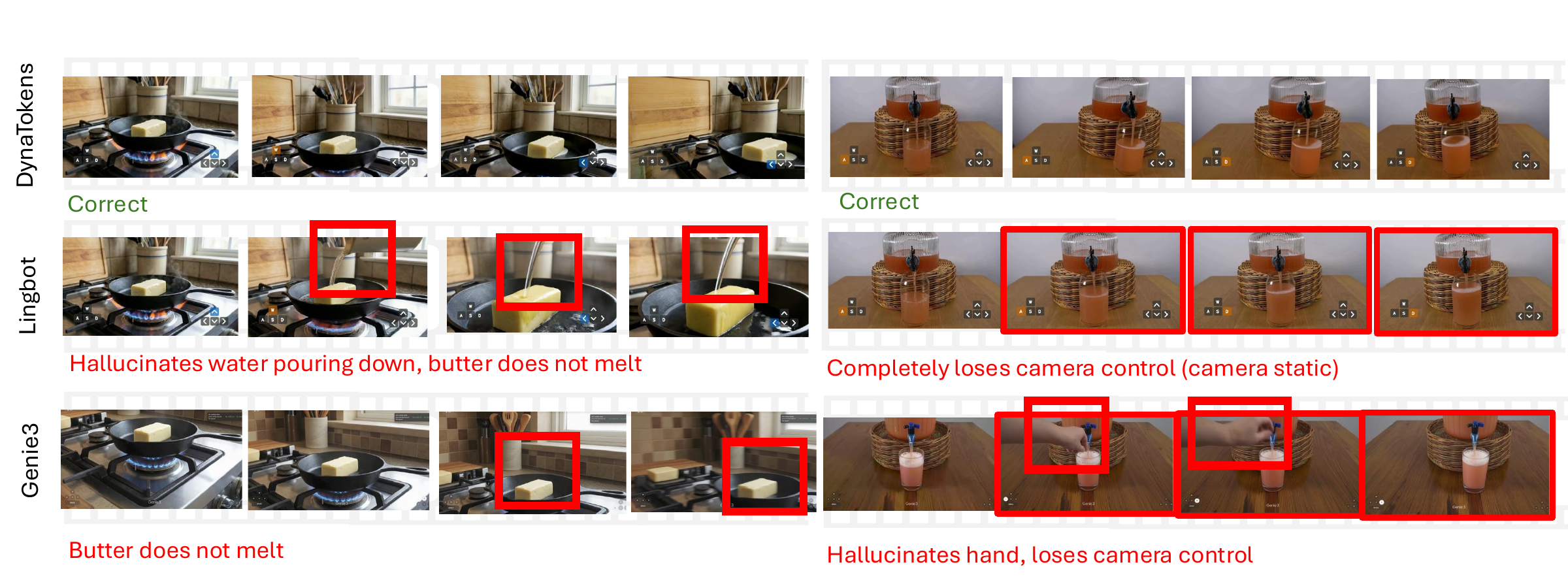}
  \caption{Additional qualitative examples for physical dynamics.}
  \label{fig:physappendix}
\end{figure}

\begin{figure}[h]
  \centering
  \includegraphics[width=\textwidth]{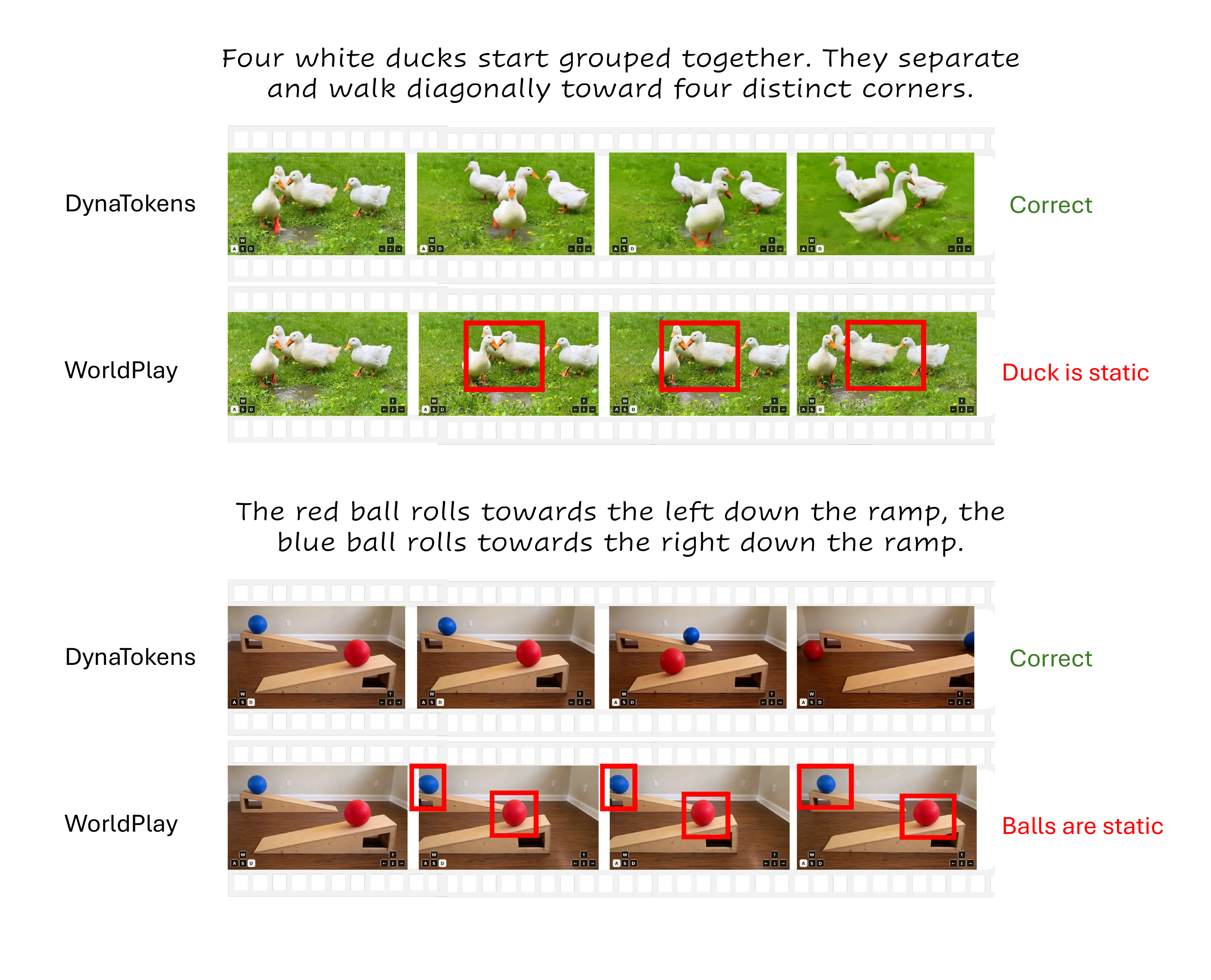}
  \caption{DynaTokens can enable dynamics of multiple objects.}
  \label{fig:multiple_objects}
\end{figure}

Fig.~\ref{fig:baselines_appendix1} \& \ref{fig:baselines_appendix2} \& \ref{fig:baselines_appendix3} \& \ref{fig:baselines_appendix4} show additional qualitative comparisons between DynaTokens and current state-of-the-art camera-controlled models.
Fig. \ref{fig:physappendix} shows additional examples of physical dynamics, and Fig. \ref{fig:multiple_objects} shows that DynaTokens can enable dynamics of multiple objects.
Fig.~\ref{fig:ttt_appendix1} \& \ref{fig:ttt_appendix2} show additional comparisons between DynaTokens and other test-time training methods. For interactive video visualization, please visit our project webpage \url{https://glab-caltech.github.io/dynatokens/}. 

\section{Evaluation: Standard Errors}
\begin{table}[h]
\begin{center}

\begin{tabular}
{l|ccc|cc}
& \multicolumn{3}{c|}{VBench2} & \multicolumn{2}{c}{WorldScore} \\
  & DSR & MOU & Cam & MA & Cam \\
\midrule
Lyra2 & 0.54 (0.071) & 0.17 (0.042)& 0.87 (0.017) & 1.11 (0.226)& \textbf{0.70 (0.015)} \\
Lingbot & 0.68 (0.071)& 0.25 (0.058)& 0.82 (0.017)& 4.37 (0.605)& 0.64 (0.023)\\
WorldPlay & 0.56 (0.070)& 0.07 (0.032)& \textbf{0.88 (0.009)} & 1.95 (0.273)& 0.69 (0.013)\\
Hydra & 0.51 (0.066)& 0.18 (0.050)& 0.61 (0.019)& 2.51 (0.520)& 0.56 (0.029)\\
LiveWorld & 0.44 (0.061) & 0.35 (0.062)& 0.82 (0.016)& 4.49 (0.817) & \textbf{0.70 (0.013)} \\
DynaTokens & \textbf{0.96 (0.030)} & \textbf{0.69 (0.061)} & 0.87 (0.011) & \textbf{5.33 (0.531)} & 0.69 (0.015)\\
\bottomrule
\end{tabular}

\end{center}
\caption{Comparison of DynaTokens with camera-controlled video models on VBench2 and WorldScore with standard errors across test samples. DynaTokens' advantage is significant.}
\label{table:baselines_se}
\end{table}
\begin{table}[h]
\begin{center}
\begin{tabular}
{l|ccc|cc}
& \multicolumn{3}{c|}{VBench2} & \multicolumn{2}{c}{WorldScore} \\
  & DSR & MOU & Cam & MA & Cam \\
\midrule
Dynatoken & \textbf{1.00 (0.000)} & \textbf{0.72 (0.082)} & \textbf{0.88 (0.011)} & \textbf{4.46 (0.538)} & \textbf{0.67 (0.060)} \\
LoRA & 0.61 (0.113) & 0.22 (0.101)& 0.83 (0.008)& 2.79 (0.523) & 0.56 (0.062)\\
LoRA-no PRoPE & 0.41 (0.123) & 0.06 (0.056)& 0.84 (0.008)& 3.62 (0.337) & \textbf{0.67 (0.056)} \\
Finetuning & 0.59 (0.123) & 0.67 (0.114) & 0.83 (0.021) & 2.66 (0.358)& 0.54 (0.064)\\
TTT Layer & 0.41(0.110) & 0.11 (0.076) & \textbf{0.88 (0.007)} & 3.79 (0.384)& 0.66 (0.061) \\
\bottomrule
\end{tabular}
\end{center}
\caption{Comparison of DynaTokens with other test-time training methods, including LoRA, block finetuning, and TTT layer, with standard errors across evaluation examples. DynaTokens' advantage is significant.}
\label{table:ablation_method_se}
\end{table}

Table~\ref{table:baselines_se} and Table~\ref{table:ablation_method_se} report standard errors for each method across the test set. DynaTokens' advantage is significant both compared to baseline world models and compared to alternative test-time training methods.

\section{Additional Comparison with VPT/APT}
\begin{wraptable}{r}{0.5\linewidth} 
\vspace{-3mm}
\begin{center}
\resizebox{1.0\linewidth}{!}{
\begin{tabular}
{l|ccc|cc}
& \multicolumn{3}{c|}{VBench} & \multicolumn{2}{c}{WorldScore} \\
  & DSR & MOU & Cam & MA & Cam \\
\midrule
DynaTokens & \textbf{1.00} & \textbf{0.75} & \textbf{0.86} & \textbf{5.15} & \textbf{0.69} \\
VPT/APT & 0.50 & 0.17 & 0.84 & 2.35 & 0.67 \\
\bottomrule
\end{tabular}
} 
\end{center}
\caption{Comparison of DynaTokens and VPT/APT. Naive prefix tuning methods cannot disentangle dynamics from camera control, and struggles to learn dynamics effectively.}
\label{table:vptapt}
\end{wraptable}

We perform additional comparison with prefix-tuning methods, Visual Prompt Tuning (VPT)~\citep{jia2022vpt} (and its video version APT~\cite{Bandara2024apt}) on a subset of evaluation scenes. The implementation of VPT/APT in the camera-controlled video generation setting is the same. When trained on the same data, VPT/APT struggles to learn dynamics effectively, stays close to the static settings for many (especially unseen) trajectories, and sometimes generates artifacts such as duplicating objects. Quantitatively, DynaTokens shows a significant advantage in Table \ref{table:vptapt}.

The comparison with VPT and APT illustrates an important design aspect of DynaTokens. While VPT and APT were originally designed for recognition (VPT for images, and APT a more efficient version for videos), applying them to camera-controlled video generation means inserting trainable “data tokens” alongside the original data tokens. While DynaTokens also inserts new tokens, the mechanism of token interaction with the base model is very different. Unlike VPT/APT, DynaTokens does not participate in the global self attention alongside the data tokens, where camera control takes place. As shown in Fig.~\ref{fig:dynatoken}(a) in the main paper, the camera conditioning happens via PRoPE in the self-attention. VPT/APT-style methods will interfere with camera conditioning (the added tokens are entangled in the camera and data path), thus failing to disentangle object movement from camera movement. This entanglement is exactly the source of issues with camera-controlled video models and the motivation behind DynaTokens.

\section{Additional Analysis of DynaTokens' Attention Map Localization}
\label{sec:attention_localization}

\begin{table}[t!]
\begin{center}
\begin{tabular}
{l|cc}
  & IoU & IoU-d1 \\
\midrule
DynaTokens & \textbf{0.342} & \textbf{0.511} \\
LoRA & 0.006 & 0.017 \\
Text & 0.011 & 0.024\\
Chance & 0.010 & 0.022 \\
\bottomrule
\end{tabular}
\end{center}
\caption{IoU and IoU-d1 (relaxed to count all distnace $\leq$1 neighbors as correct) of DynaTokens attention map, attention map delta before and after LoRA, the attention map from text keywords corresponding to the moving object (such as the word ``dog''), as well as a random attention mask pattern. DynaTokens has a significantly higher overlap with the SAM mask of the moving object, while LoRA and text perform similarly to random chance.}
\label{table:sam}
\vspace{-7mm}
\end{table}
We quantify DynaTokens' attention localization by measuring the overlap between attention maps with SAM~\cite{ravi2025sam} masks via IoU and IoU-d1 (relaxing the IoU by counting all distance $\leq$1 neighbors as correct). We note that while the attention map illustrates DynaTokens’ localization to the motion, its purpose is not to segment the exact object geometry in video frames. Additionally, DynaTokens work in latent patches which are coarse spatially and temporally (a latent patch is 16x16 pixels across 4 physical frames), and thus we expect coarse overlap rather than accurate, high-IoU boundary alignment. We compare with the attention map delta caused by LoRA, as well as the attention map from the text keyword of the moving object. We additionally show the expected IoU of a random attention mask pattern, denoted by “Chance”. As shown in Table~\ref{table:sam}, DynaTokens has a significantly higher overlap with the SAM mask of the moving object, while LoRA and text perform similarly to random chance.

We further calculate Ripley’s L statistic which measures whether points in 2D have a clustered distribution pattern (higher means more clustered). DynaTokens scores 5.95, LoRA 4.44, text keyword 2.17 (L(3)-3), which shows a clear lead by DynaTokens.

\begin{figure}[h]
  \centering
  \includegraphics[width=0.8\textwidth]{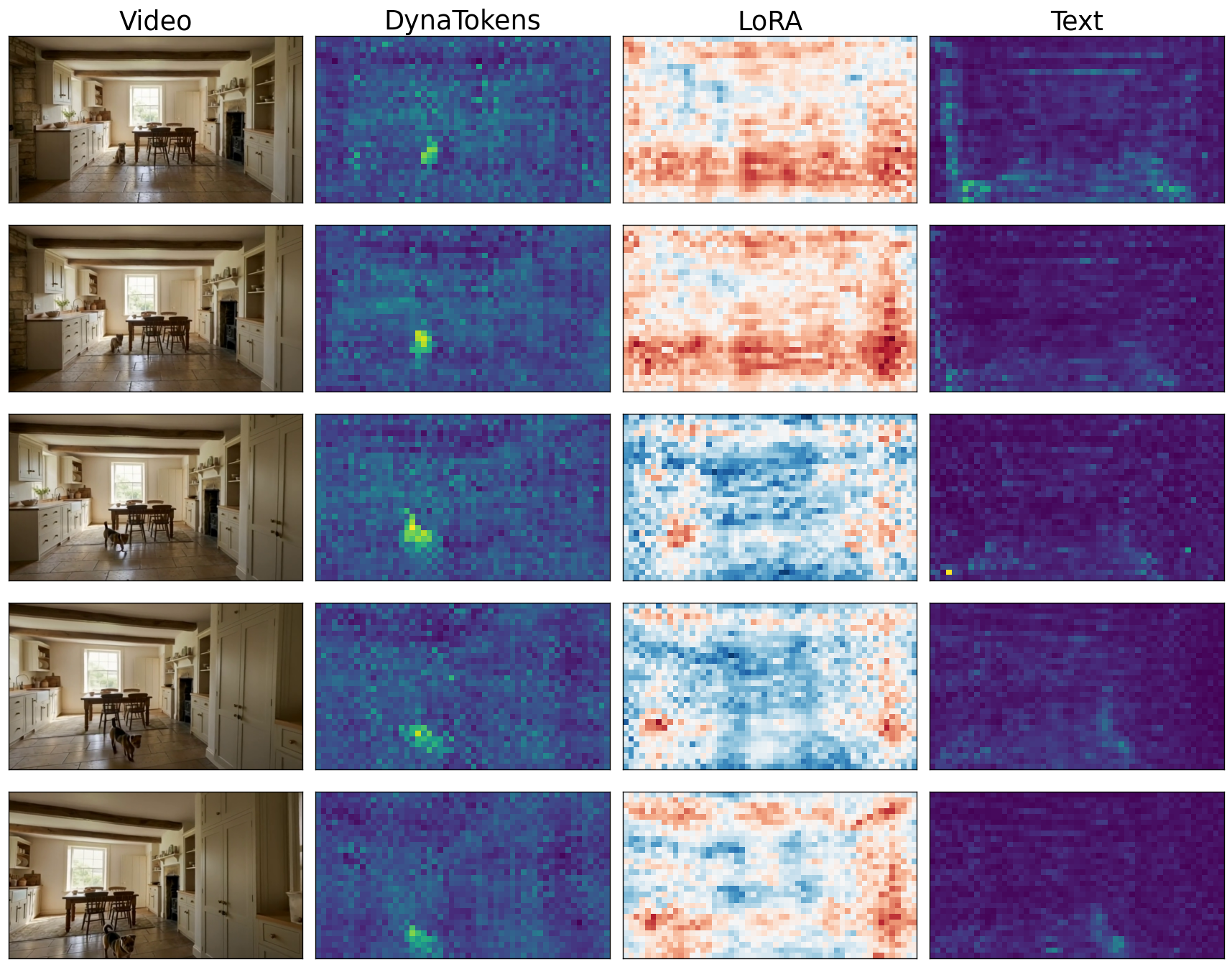}
  \caption{Attention map visualization of DynaTokens, LoRA, and the text keyword of the moving object (dog).}
  \label{fig:attentionmap_dog}
\end{figure}

\begin{figure}[h]
  \centering
  \includegraphics[width=0.8\textwidth]{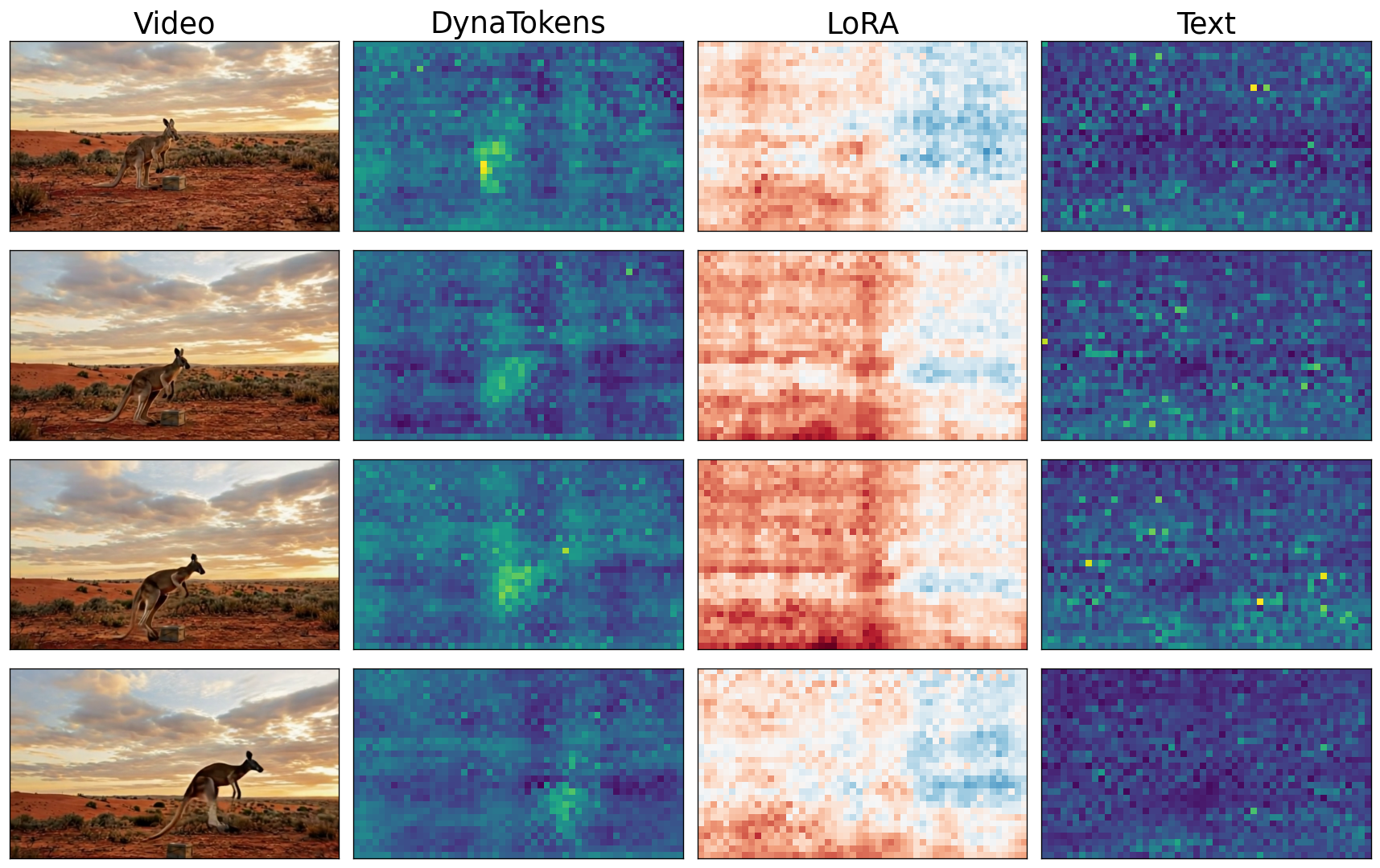}
  \caption{Attention map visualization of DynaTokens, LoRA, and the text keyword of the moving object (kangaroo).}
  \label{fig:attentionmap_kangaroo}
\end{figure}

\begin{figure}[h]
  \centering
  \includegraphics[width=0.8\textwidth]{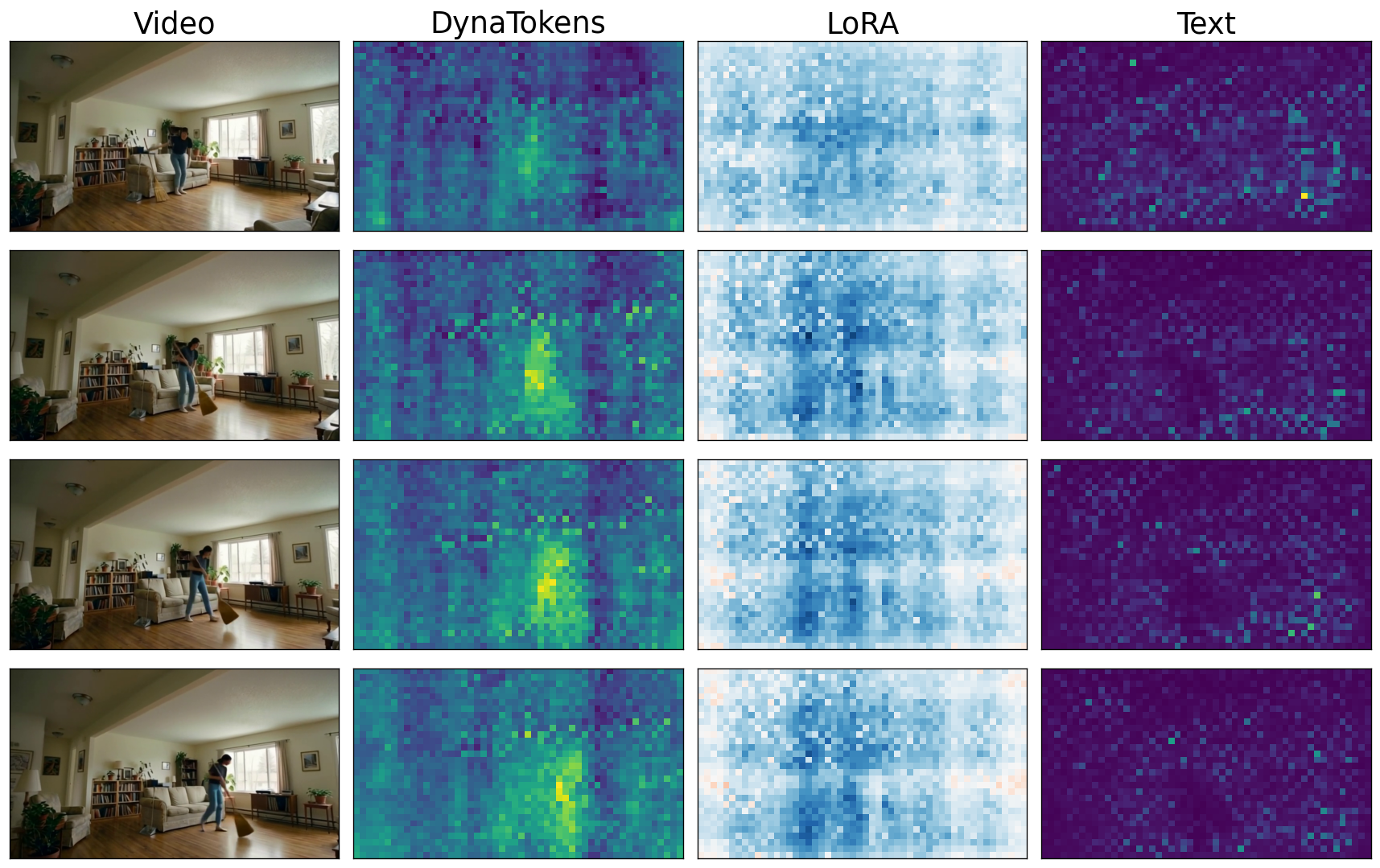}
  \caption{Attention map visualization of DynaTokens, LoRA, and the text keyword of the moving object (person).}
  \label{fig:attentionmap_person}
\end{figure}

\begin{figure}[h]
  \centering
  \includegraphics[width=0.8\textwidth]{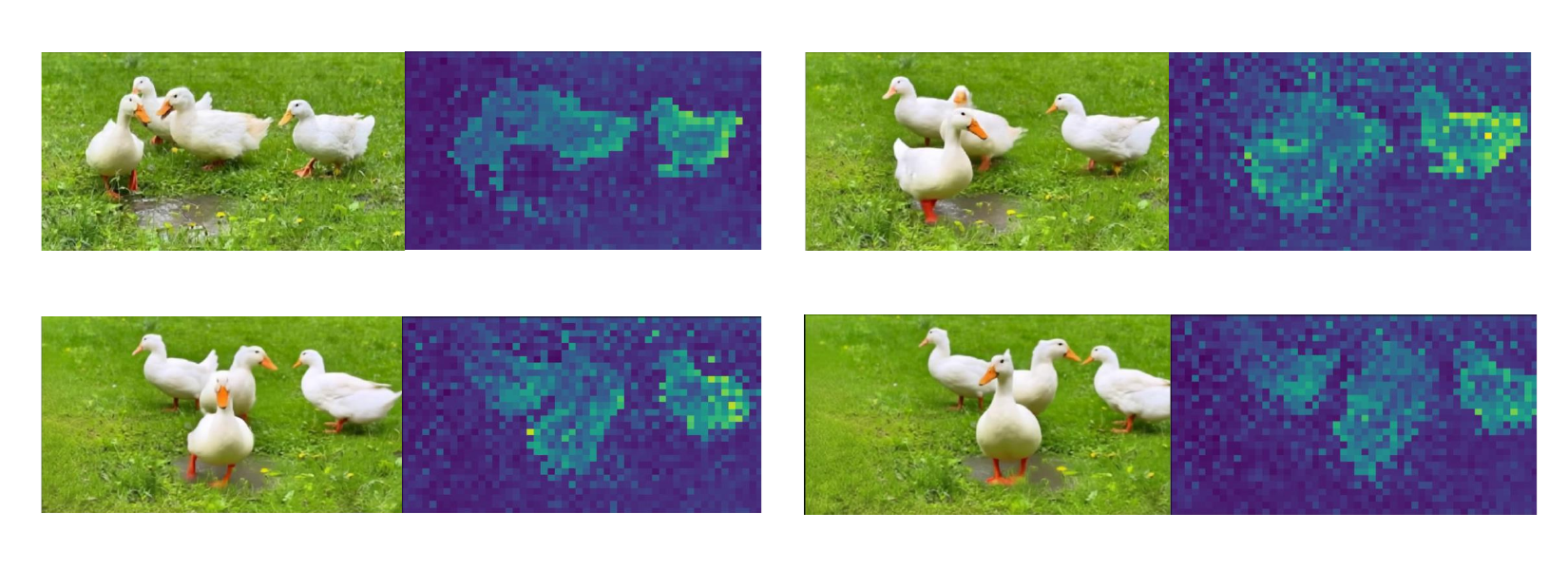}
  \caption{Attention map visualization of DynaTokens for a complex scene with four moving objects (four ducks moving in different directions).}
  \label{fig:attentionmap_ducks}
\end{figure}

Fig.~\ref{fig:attentionmap_dog} \& \ref{fig:attentionmap_kangaroo} \& \ref{fig:attentionmap_person} show additional attention map visualizations with DynaTokens and compares them to LoRA and text-based attention maps.
Fig.~\ref{fig:attentionmap_ducks} further shows attention map localization for a challenging scenario where 4 objects are moving (4 ducks moving in different directions). These visualizations further support our localized dynamics hypothesis that motivates our DynaTokens design. 

\section{Robustness to Training Trajectories}

\begin{wraptable}{r}{0.5\linewidth} 
\vspace{-12mm}
\begin{center}
\resizebox{\linewidth}{!}{
\begin{tabular}
{l|ccc|cc}
\# trajectories & \multicolumn{3}{c|}{VBench} & \multicolumn{2}{c}{WorldScore} \\
  & DSR & MOU & Cam & MA & Cam \\
\midrule
3 & 0.67 & 0.50 & 0.80 & 4.02 & 0.67 \\
6 & 0.75 & 0.67 & 0.80 & 4.68 & 0.68 \\
12 & 1.00 & 0.75 & 0.86 & 4.81 & 0.68 \\
15 & 1.00 & 0.75 & 0.87 & 5.15 & 0.68 \\
\bottomrule
\end{tabular}
} 
\end{center}
\vspace{-2mm}
\caption{Studying the effect of the number of training trajectories. Even when training on as few as 3 trajectories, DynaTokens shows a considerable advantage on MOU (0.50 vs. 0.35 for the best baseline, LiveWorld).}
\vspace{-5mm}
\label{table:training_quantity}
\end{wraptable}
We perform additional ablations on a subset of evaluation scenes to evaluate the robustness of DynaTokens to the quantity and quality of training trajectories.
In Table~\ref{table:training_quantity}, we perform an additional study by training on 3/6/12 trajectories on a subset of 30 evaluation videos. Notably, even when training on as few as 3 trajectories, DynaTokens shows a considerable advantage on MOU (0.50 vs. 0.35 for the best baseline, LiveWorld), which is a positive result. Other metrics (DSR, MOU, MA) beat baselines even at 6 trajectories. This shows that DynaTokens is effective even with a small number of training trajectories.

\begin{wraptable}{r}{0.5\linewidth}
\vspace{-8mm}
\begin{center}
\resizebox{\linewidth}{!}{
\begin{tabular}
{l|ccc|cc}
\# artifact swap & \multicolumn{3}{c|}{VBench} & \multicolumn{2}{c}{WorldScore} \\
  & DSR & MOU & Cam & MA & Cam \\
\midrule
0 (original) & 1.00 & 0.75 & 0.87 & 5.15 & 0.68 \\
1 (7\%) & 1.00 & 0.75 & 0.86 & 4.99 & 0.66 \\
5 (33\%) & 1.00 & 0.67 & 0.86 & 4.50 & 0.66 \\
\bottomrule
\end{tabular}
} 
\end{center}
\vspace{-2mm}
\caption{Studying the effect of the quality of training trajectories by swapping k good trajectories with k trajectories with artifacts. Even with 33\% trajectories swapped to corrupted, DynaTokens' performance remains high.}
\vspace{-7mm}
\label{table:training_quality}
\end{wraptable}
In Table~\ref{table:training_quality}, we perform an additional ``swapping” analysis on a subset of 30 evaluation videos: we swap up to 33\% of good training trajectories with corrupted trajectories (containing background inconsistencies or sudden motion changes). We observe DynaTokens’ performance remains high, above baselines even with 33\% of training trajectories corrupted by artifacts. This is likely due to the DynaTokens formulation retaining the base model prior of coherent video generation.

\section{Benchmark Evaluation Details}
\label{sec:metrics_details}

For VBench2 evaluation, we use the ``Dynamic Spatial Relations'' and ``Motion Order Understanding'' categories. We use VBench's original VLM evaluation framework with minor modifications to accommodate camera movement, since the original framework is designed for static camera. For example, VBench2 determines object spatial relation by the first and last frame. Since camera-controlled models might occasionally turn the camera away from the main object (\eg camera turns left while object moves right and out of frame), we add an ``object presence'' detection and only apply this check when the main object is present in the last frame. We also remove the questions solely based on the first frame, since all methods we test take the same first frame as input. We use Gemini 2.5 Flash instead of Llava for better spatial relation understanding. To avoid Gemini bias, we additionally use GPT-5 for verification, which yields very similar results: the mean difference is 0.025, which is less than 10\% of the observed advantage of DynaTokens over the best baseline (Lingbot) on dynamic spatial relations. For methods that generate highly distorted or unrecognizable subjects, we add ``recognizable and not distorted'' to the evaluation prompt. We use a state-of-the-art camera estimation method, ViPE~\citep{huang2025vipe} to estimate camera, and calculate camera score based on a geometric mean of rotation and translation errors, where rotation error is denoted by geodesic distance, similar to WorldScore~\citep{duan2025worldscore}. Since the translation is dependent on the unit for each model, we apply scaling to correct for the unit conversion.

The WorldScore benchmark's dynamic evaluation is largely based on optical flow, and assumes a static camera (thus low background flow). To adapt it to the moving camera setting, we adapt the Motion Accuracy and Camera Control metrics. Two other metrics, Motion Magnitude and Motion Smoothness, are both dominated by camera-induced flow rather than scene dynamics and are thus not suitable for this task. Motion Accuracy uses SEA-RAFT~\citep{searaft2024} dense flow and SAM2~\citep{ravi2024sam2segmentimages} object masks to score whether the prompted subject moves more than its surroundings; we (i) set the object-region flow to zero when SAM2 fails to localize the subject, so lost or distorted subjects are penalized rather than collapsing into a whole-frame statistic, and (ii) replace the background maximum with the background \emph{median}, since under camera motion the background flow distribution is heavy-tailed and its maximum is unstable. Camera Control recovers the trajectory via DROID-SLAM~\citep{teed2022droidslamdeepvisualslam} and scores the clipped geometric mean of rotation geodesic error and scale-aligned translation $\ell_2$ error, with $C^{*} = \arg\min_C \|t_{\text{gt}} - C\, t_{\text{pred}}\|_2$; we constrain $C \in [c_{\min}, c_{\max}]$ to avoid the loophole where under-translating prediction is rescaled to match the ground-truth magnitude.

For evaluation of the baseline camera-controlled models, we apply conversion from action string (which WorldPlay takes as input) to camera matrices for Lingbot, Lyra2, LiveWorld and Hydra. Since Hydra only takes video input, and our setting does not have history, we repeat the initial frame as the input video.

\section{Test-Time Training Details}

\paragraph{Data Curation.}
\label{sec:curation_appendix}

We generate video samples with dynamics to train dynamic tokens. Given an initial frame $I_0$, text prompt $\tau$, and a target camera trajectory $\pi$ partitioned into $N$ segments, $\pi_{1{:}1},\ldots,\pi_{1{:}N}$, we construct an imperfect video supervision $(I_0, \tau, \pi) \mapsto V$:

\textit{(i) Dynamics synthesis.} A VLM (Gemini 3.1 Pro) first enhances $\tau$ by making the dynamics more explicit. Then, an image-to-video model conditioned on $I_0$ and the enhanced caption is used to produce a static-camera clip $V_\mathrm{dyn}$. The same VLM then scores $V_\mathrm{dyn}$ on physical plausibility and camera staticity, refining the caption until passes.

\textit{(ii) Camera projection.} We sample $N{+}1$ uniformly spaced state
frames $\{S_0,\ldots,S_N\}$ from $V_\mathrm{dyn}$. For each $i$, we run HY-WorldPlay on
$(S_i,\, \tau_\varnothing,\, \pi_{1{:}i})$ with the neutral prompt
$\tau_\varnothing{=}$``\texttt{A static scene.}'', recovering a camera-aligned view $\hat{S}_i$ at the boundary of
$\pi_{1{:}i}$.  For instance, with $\pi{=}$``\texttt{forward-6, right-7, forward-6}'' and $N{=}3$, we run HY-WorldPlay three times: on $S_1$ with prefix $\pi_{1{:}1}{=}$``\texttt{forward-6}'', on $S_2$ with $\pi_{1{:}2}{=}$``\texttt{forward-6, right-7}'', and on $S_3$ with $\pi_{1{:}3}{=}\pi$. This yields $\hat{S}_1,\hat{S}_2,\hat{S}_3$ that each capture the scene state $S_i$ as observed from the camera position reached after executing $\pi_{1{:}i}$.

\textit{(iii) Interpolation and stitching.} Consecutive keyframes
$(\hat{S}_{i-1}, \hat{S}_i)$ are bridged by a dual-endpoint image-to-video interpolator, Kling o3~\citep{klingteam2025klingomnitechnicalreport}, guided by a transition caption that the VLM writes from the two endpoints and the camera-action label. The per-segment clips are concatenated into the final video $V$, which is passed to a VLM for review.

\begin{figure}[h]
  \centering
  \includegraphics[width=\textwidth]{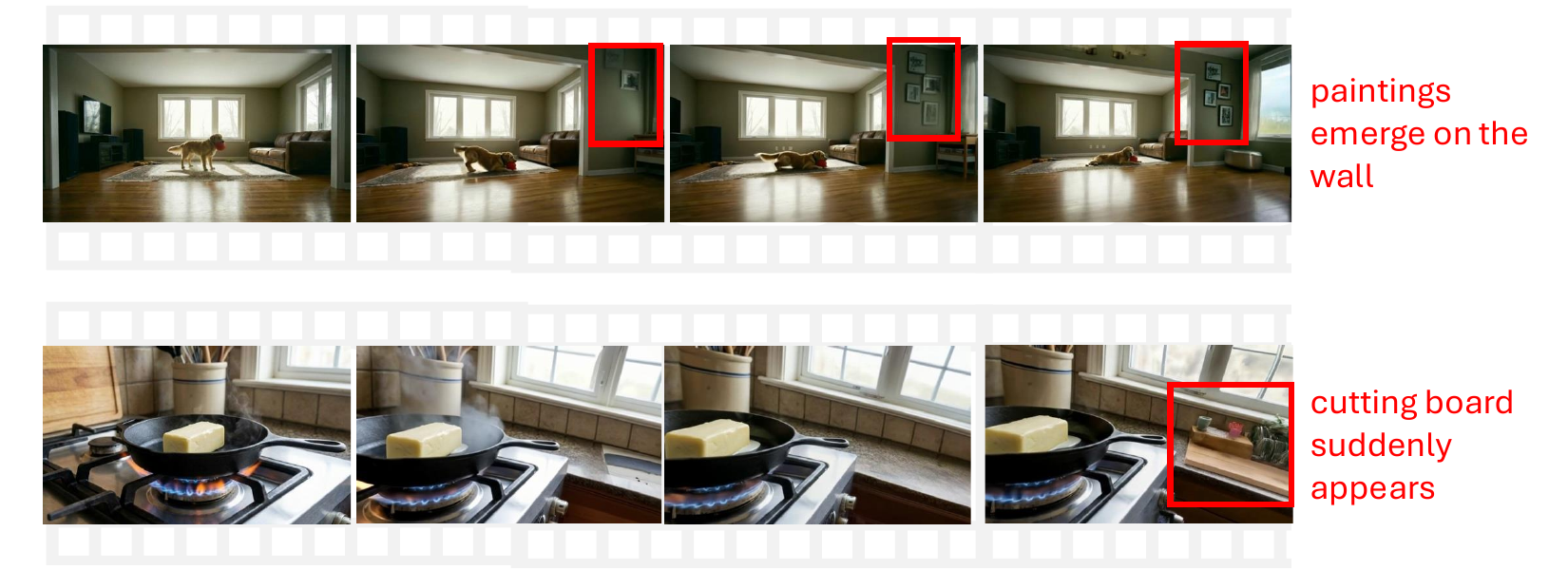}
  \caption{Examples of curated training examples. Curating one sample takes 20 GPU minutes (A100) and cost \$1.52 in API calls (Kling and Gemini). The curated data contains correct dynamics but may include background inconsistencies. Our test-time training approach resolves such inconsistencies by leveraging model priors.}
  \label{fig:data_curation}
\end{figure}

It is important to note that, although our curated videos contain both scene and camera motion, stitching can introduce artifacts in background appearance and motion continuity. For example, the scene layout or object appearance may change across frames, and temporal discontinuities may arise. Qualitative examples can be found in Fig.~\ref{fig:data_curation}. However, since the base model remains frozen and dynamic tokens contribute only through localized, additive cross-attention, this imperfect supervision can teach scene dynamics without corrupting the pretrained prior. Model inference calls take 16 minutes on 8 A100 GPUs (with shared prefix skipping), and API call for Gemini and Kling cost \$1.52 for all training trajectories. Our DynaTokens design can train on few trajectories, enabling camera-controlled models to produce correct dynamics for new, unseen camera trajectories.

We show examples of curated data below. They contain background inconsistencies since each keyframe's projection might not ``hallucinate'' unseen space in the same way.

\paragraph{Training.}
We train $K=16$ dynamic tokens of dimension $m=64$ in our experiments, for videos ranging from 61 to 93 frames. We use the Muon~\cite{jordan2024muon} optimizer with learning rate 3e-3 with weight decay and L2 regularization. The training memory is around 22GB on 4 GPUs, for which we use 40GB or 80GB A100s. For context noising, we add high noise (step [500, 985)) to the context chunks. Training takes 80 minutes for 15 trajectories (total 900 frames). To put this in context, just doing inference using Lyra2, a baseline video model, on our 6 evaluation trajectories, takes 66 A100 minutes.  This suggests the test-time adaptation, which unlocks inference on diverse unseen trajectories, is not a huge overhead compared to the per-trajectory inference cost.

\paragraph{Training limitations.}
DynaTokens is a lightweight module added to pretrained video models. Dynamic tokens enable dynamics by effectively disentangling it from camera control. However, if the base model is poor at certain dynamics, learning with dynamic tokens should not expect to fully fix these issues, such as complex physics.

\section{Additional Analysis on LoRA's Failure}
\label{sec:lora_appendix}

We provide more detailed derivation on LoRA's effect on PRoPE:
\begin{equation}
q_i = (W_Q + \Delta W_Q) h_i, \qquad
k_j = (W_K + \Delta W_K) h_j
\end{equation}

\begin{equation}
\tilde{q}_i = D(c_i)\, q_i, \qquad
\tilde{k}_j = D(c_j)\, k_j,
\end{equation}
\[
\begin{aligned}
& s_{ij}
=
\big\langle D(c_i) W_Q h_i,\; D(c_j) W_K h_j \big\rangle 
+
\big\langle D(c_i) \Delta W_Q h_i,\; D(c_j) W_K h_j \big\rangle \\
& +
\big\langle D(c_i) W_Q h_i,\; D(c_j) \Delta W_K h_j \big\rangle
+
\big\langle D(c_i) \Delta W_Q h_i,\; D(c_j) \Delta W_K h_j \big\rangle
\end{aligned}
\]

Only the first term corresponds to the original PRoPE value.

We also provide more analysis on why LoRA-No PRoPE fails.
While it does not significantly reduce the number of free parameters compared to normal LoRA, this might be causing great discrepancy between the PRoPE vs. non-PRoPE stream (which was designed to be symmetric in the original architecture), thus hindering learning. This shows that LoRA, with its global entanglement across spatial patches, struggles to only learn dynamics without affecting camera control. Dynamic tokens, on the other hand, minimizes the effect on PRoPE by being local - unaffected patches' keys and values stay mostly stable.

\section{Societal Impact}
We improve generative video models. We acknowledge that generative models can be used in ways that raise ethical concerns, including the creation of misleading synthetic media, amplification of societal biases present in training data, and potential misuse for surveillance or harmful automation. Our work is intended solely for scientific research purposes and focuses on advancing the understanding and capability of generative modeling systems. We do not release systems or artifacts designed for deceptive or malicious use, and we encourage responsible deployment practices, including transparency about generated content, careful dataset curation, and evaluation for fairness and safety.

\begin{figure}[h]
  \centering
  \includegraphics[width=\textwidth]{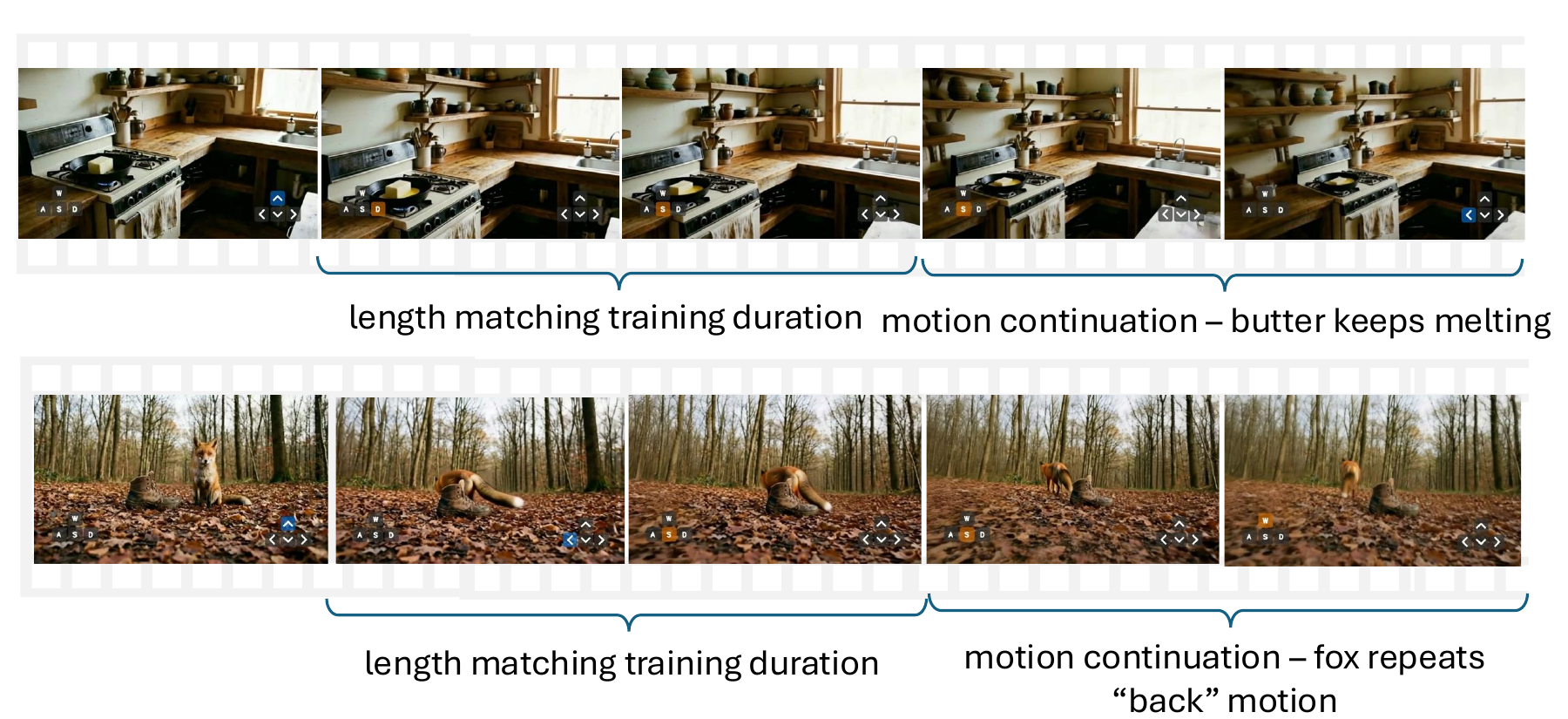}
  \caption{Evaluation on longer videos.}
  \label{fig:longvideo_appendix}
\end{figure}
\clearpage


\end{document}